\documentclass{article}

\usepackage{microtype}
\usepackage{graphicx}
\usepackage{subcaption}
\usepackage{booktabs} 
\usepackage{multirow}

\usepackage{hyperref}

\usepackage[preprint]{icml2026}

\usepackage{amsmath}
\usepackage{amssymb}
\usepackage{mathtools}
\usepackage{amsthm}

\usepackage[capitalize,noabbrev]{cleveref}

\theoremstyle{plain}
\newtheorem{theorem}{Theorem}[section]
\newtheorem{proposition}[theorem]{Proposition}
\newtheorem{lemma}[theorem]{Lemma}
\newtheorem{corollary}[theorem]{Corollary}
\theoremstyle{definition}

\newtheorem{assumption}[theorem]{Assumption}
\theoremstyle{remark}

\usepackage[textsize=tiny]{todonotes}

\icmltitlerunning{Unsupervised Text Segmentation via Kernel Change-Point Detection on Sentence Embeddings}

\begin{document}

\twocolumn[
  \icmltitle{Unsupervised Text Segmentation via Kernel Change-Point Detection on Sentence Embeddings}



  \icmlsetsymbol{equal}{*}

  \begin{icmlauthorlist}
    \icmlauthor{Mumin~Jia}{equal,yyy}
    \icmlauthor{Jairo~Diaz-Rodriguez}{equal,yyy}
  \end{icmlauthorlist}

  \icmlaffiliation{yyy}{Department of Mathematics and Statistics, York University, Toronto, Canada}

  \icmlcorrespondingauthor{Mumin~Jia}{amyjia@yorku.ca}
  \icmlcorrespondingauthor{Jairo~Diaz-Rodriguez}{jdiazrod@yorku.ca}
  \icmlkeywords{Kernel Change-Point Detection, Text Segmentation, Unsupervised Machine Learning}

  \vskip 0.3in
]



\printAffiliationsAndNotice{\icmlEqualContribution}

\begin{abstract}
Unsupervised text segmentation is crucial because boundary labels are expensive, subjective, and often fail to transfer across domains and granularity choices. We propose Embed-KCPD, a training-free method that represents sentences as embedding vectors and estimates boundaries by minimizing a penalized KCPD objective. Beyond the algorithmic instantiation, we develop, to our knowledge, the first dependence-aware theory for KCPD under $m$-dependent sequences, a finite-memory abstraction of short-range dependence common in language. We prove an oracle inequality for the population penalized risk and a localization guarantee showing that each true change point is recovered within a window that is small relative to segment length. To connect theory to practice, we introduce an LLM-based simulation framework that generates synthetic documents with controlled finite-memory dependence and known boundaries, validating the predicted scaling behavior. Across standard segmentation benchmarks, Embed-KCPD often outperforms strong unsupervised baselines. A case study on Taylor Swift's tweets illustrates that Embed-KCPD combines strong theoretical guarantees, simulated reliability, and practical effectiveness for text segmentation.
\end{abstract}

\section{Introduction}

Segmenting a document into coherent topical units is a core subroutine in many NLP and IR systems.
Reliable boundaries improve retrieval, summarization, question answering, discourse analysis, and downstream modeling \citep{textsegmentation-ir, topicsegmentation-ir, efficient-ir, cho-etal-2022-toward}.
Despite this importance, text segmentation is often a poor fit for standard supervised learning.
The ``correct'' boundary locations depend on the downstream task, the desired granularity, and the annotation protocol, which can vary substantially across corpora.
Labels are therefore expensive to obtain, difficult to standardize, and may not transfer cleanly across domains.
This makes \emph{unsupervised} segmentation particularly valuable in practice: a method that can be deployed without training labels and remains robust across datasets is often more useful than a narrowly optimized supervised model.

Change-point detection (CPD) provides a natural statistical lens for unsupervised segmentation: boundaries correspond to indices where the data-generating distribution changes. Classical offline CPD methods come with strong guarantees, but these often rest on restrictive assumptions such as Gaussianity, independence, or homoscedasticity \citep{Basseville1993, dynamic2002, Killick01122012}, which can be brittle for high-dimensional text representations. Kernel change-point detection (KCPD) relaxes much of this structure by comparing distributions through RKHS embeddings, enabling detection of rich distributional shifts without explicit density estimation \citep{Harchaoui2007RetrospectiveMC, arlot-2019}. This makes KCPD a natural fit for embedding-based segmentation, where modern sentence encoders can reveal semantic changes even when lexical cues are weak. At the same time, deploying KCPD in text exposes a key theoretical limitation: most existing analyses assume independent observations \citep{garreau-consistenkcpd}, while language exhibits ubiquitous short-range dependence because adjacent units share context, discourse structure, and lexical overlap. This gap motivates dependence-aware guarantees tailored to sequential text.

This paper introduces \textbf{Embed-KCPD}, a modular, training-free method for unsupervised text segmentation that combines pretrained sentence embeddings with kernel change-point detection, and provides statistical guarantees for the resulting estimator.
Given a sequence of text units $X_1,\dots,X_T$, we compute embeddings $Y_t = f(X_t) \in \mathbb{R}^d$ using a fixed encoder $f$.
We then estimate change points by minimizing a penalized KCPD objective.
KCPD is attractive for this setting because it detects general distributional changes, not only mean shifts, while remaining nonparametric and compatible with high-dimensional representations \citep{Harchaoui2007RetrospectiveMC, arlot-2019}.
Moreover, the penalized objective can be optimized exactly with dynamic programming and efficiently with pruning (PELT), which makes the method practical for long documents \citep{Killick01122012}.
The resulting pipeline cleanly decouples representation learning from statistical segmentation, so improvements in sentence encoders can be used immediately without retraining the segmenter.

Beyond proposing a practical method, our goal is to provide a principled foundation for dependent text sequences.
To bridge the gap between i.i.d. theory and sequential language, we develop, to our knowledge, the first guarantees for penalized KCPD under \textbf{$m$-dependence}, a tractable abstraction of finite-memory dependence. While natural language is not literally 
$m$-dependent, this finite-range model offers a clean first approximation to short-range contextual dependence and enables sharp analysis.
Under this dependency assumption, we prove an oracle inequality for the population penalized risk and we establish a localization result showing that true change points are recovered within a window whose size is small relative to the segment lengths, yielding vanishing relative error as $T$ grows.

We connect these results to practice in two complementary ways.
First, we introduce a controlled simulation framework that uses an LLM to generate synthetic documents with known change points and explicit finite-memory dependence, enabling stress tests that mirror realistic sequential text while retaining ground truth.
Second, we provide a systematic empirical study of Embed-KCPD for text segmentation across standard benchmarks and multiple modern encoders.
Across datasets, Embed-KCPD is competitive with established unsupervised baselines and often improves standard segmentation metrics.
A case study on a long-running tweet stream illustrates that the discovered segments align with interpretable thematic phases and can support downstream exploratory analysis.

\paragraph{Contributions.}
Our main contributions are: (i) dependence-aware analysis of penalized KCPD under \textbf{$m$-dependence}, including an oracle inequality and a change-point localization guarantee;
(ii) \textbf{Embed-KCPD}, a simple, modular, training-free pipeline for \textbf{unsupervised} text segmentation that applies offline KCPD to pretrained sentence embeddings;
(iii) an LLM-based simulation framework for generating short range dependent text with known boundaries, used to validate the predicted scaling behavior; and
(iv) an extensive empirical evaluation on diverse segmentation benchmarks showing that Embed-KCPD is a strong and practical unsupervised baseline.


\section{Related Work}

\textbf{Change-point detection methods}. Classical algorithms include Binary Segmentation \citep{scott1974}, dynamic programming \citep{dynamic2002}, and the Pruned Exact Linear Time (PELT) method \citep{Killick01122012}, which offer consistency guarantees under parametric cost functions. Nonparametric approaches relax such assumptions using rank or divergence measures \citep{Aminikhanghahi2017}, while kernel methods embed data into reproducing kernel Hilbert spaces \citep{harchaoui2007-KCP-analysis}. Recent work explores online and streaming algorithms for real-time detection \citep{ferrari-online, hushchyn2020onlineneuralnetworkschangepoint}, ensemble and statistical inference methods for more reliable boundaries \citep{leduy-optimal, shiraishi2025}, deep kernel learning for adaptive representations \citep{chang2018kernel}, and unsupervised deep frameworks \citep{truong2020selective}. 

\textbf{Theoretical results on CPD beyond independence.} Beyond independence, CPD under dependence has been studied mainly for parametric or low-dimensional settings: CUSUM/MOSUM with mixing and long-run variance or self-normalization \citep{csorgo1997limit, aue2012structural, horvath2014extensions}, econometric structural-break tests with robust covariances \citep{andrews1994tests, bai1998estimating}, variance change via ICSS \citep{inclan1994use}, and penalized-contrast methods for dependent series \citep{lavielle2000least, lavielle2005penalized}, with extensions to high-dimensional mean shifts \citep{cho2015multiple, wang2018high}. To our knowledge, we provide the first theoretical results for non-parametric kernel CPD  under $m$-dependence, aligning theory with modern embedding-based text segmentation.

\textbf{Text segmentation methods.} Early methods like TextTiling \citep{hearst-1994-multi} exploit lexical cohesion, while later probabilistic approaches, including pLSA-based segmentation \citep{brants-2002}, dynamic programming over TF–IDF similarity \citep{Fragkou-2004-Dynamic}, BayesSeg \citep{bayesseg-2008}, and LDA-based extensions \citep{riedl-biemann-2012-topictiling, du-etal-2013-topic}, model topical transitions via latent distributions. Recent techniques incorporate coherence-aware segmentation, semantic or embedding signals \citep{glavas-etal-2016-unsupervised, BERTseg, coherence-2024, yu-etal-2023-improving-long, gklezakos2024treetopic}; mainly tailored to specific applications, rather than general-purpose text segmentation. In parallel, supervised methods frame segmentation as boundary classification, from attention-based BiLSTMs \citep{Badjatiya-2018} and hierarchical BiLSTMs \citep{koshorek-etal-2018-text}, to Transformer variants using cross-segment attention \citep{lukasik-etal-2020-text} and multi-level Transformer designs \citep{CATs-2020}. On the contrary our approach is fully unsupervised text segmentation.

\section{Preliminaries and Problem}\label{sec: preliminaries}

Let $Y_1,\cdots,Y_T\in\mathbb R^d$ be an observed sequence. 
A segmentation of $1,\cdots T$ into $K+1$ contiguous blocks is determined by change points
${\boldsymbol\tau}_K = (\tau_0,\tau_1,\dots,\tau_K,\tau_{K+1})$ with
$0=\tau_0 < \tau_1 < \cdots < \tau_K < \tau_{K+1}=T$. We assume there exist true change points ${\boldsymbol\tau}_K$ such that
the distribution of $(Y_t)$ is piecewise stationary across the $K+1$ blocks.
The task is to recover both $K$ and the locations $\tau_1,\dots,\tau_K$.

Let $k:\mathbb R^d\times\mathbb R^d\to\mathbb R$ be a positive definite kernel with RKHS $\mathcal H$. The mapping function $\phi$: $\mathbb R^d\to\mathcal H$ is implicitly defined by $\phi(y_t) = k(y_t, \cdot) \in \mathcal H$.
For distributions $P,Q$, the squared maximum mean discrepancy is
$\text{MMD}^2(P,Q)=\|\mu_P-\mu_Q\|_{\mathcal H}^2$.
For data $Y_s,\dots,Y_e$, define the empirical block cost
\[
\widehat C(s,e)=\sum_{t=s}^e k(Y_t,Y_t)
- \frac{1}{e-s+1}\sum_{i=s}^e\sum_{j=s}^e k(Y_i,Y_j),
\]
with expectation $C(s,e)=\mathbb E[\widehat C(s,e)]$.
Intuitively, $\widehat C(s,e)$ measures within-block dispersion in RKHS.

\paragraph{Penalized segmentation criterion.}
For a candidate segmentation ${\boldsymbol\tau}'_{K'}$, its cost is
\[
L({\boldsymbol\tau}'_{K'})=\sum_{k=1}^{K'+1}\widehat C(\tau'_{k-1}+1,\tau'_k)
+ \beta_T K',
\]
where $\beta_T$ penalizes over-segmentation.
The kernel change point detection (KCPD) estimator is
\[
\widehat{\boldsymbol\tau}_{\widehat K}
= \arg\min_{{\boldsymbol\tau}'_{K'}} L({\boldsymbol\tau}'_{K'}).
\]
$L$ can be minimized exactly with the pruned exact linear time (PELT) algorithm, which under mild conditions has computational cost linear in the document length $T$.

\section{KCPD Under $m$-Dependence}\label{sec:theory_m_dependence}

We now derive our main theoretical results for $\widehat{\boldsymbol\tau}_{\widehat K}$ under $m$-dependent data.
Our goal is to bridge the gap between the classical i.i.d.\ analyses of kernel change-point detection and the short-range dependence that arises naturally in sequential text, where adjacent units share context, discourse structure, and lexical overlap.


The following assumptions formalize the statistical setting.

\begin{assumption}[$m$-dependence + within-block stationarity]\label{A1}
The sequence $(Y_t)_{t=1}^T$ is $m$-dependent:
$Y_t\perp Y_{t'}$ whenever $|t-t'|>m$.
Moreover, for each $k=1,\dots,K+1$, the subsequence
$\{Y_t:\tau_{k-1}<t\le\tau_k\}$ is strictly stationary with distribution $P_k$.
\end{assumption}

\begin{assumption}[kernel]\label{A2}
The kernel $k:\mathbb R^d\times\mathbb R^d\to\mathbb R$ is bounded and characteristic:
$0\le k(x,y)\le M<\infty$.
Let $\mathcal H$ denote the associated RKHS.
\end{assumption}

\begin{assumption}[detectability]\label{A3}
Let $\mu_{P_k}$ be the RKHS mean embedding of block $k$ and define
$\Delta_k^2 := \|\mu_{P_k}-\mu_{P_{k+1}}\|_{\mathcal H}^2 > 0$.
Set $\Delta_\star^2 := \min_k \Delta_k^2 > 0$.
\end{assumption}

\begin{assumption}[minimum spacing]\label{A4}
The minimal block length satisfies
$\ell_T := \min_k(\tau_k-\tau_{k-1}) \to\infty$, and
$\ell_T / \sqrt{T \log T} \to \infty$ as $T\to\infty$.
\end{assumption}

\begin{assumption}[penalty]\label{A5}
The penalty $\beta_T$ satisfies
$\beta_T \;\ge\; 16M\sqrt{2(8m+5)T\log T} + 2M(1+6m),
\qquad
\beta_T = O(\sqrt{T\log T}).
$
\end{assumption}

\begin{assumption}[Admissible Segmentation]\label{A6} The estimator $\widehat{\boldsymbol\tau}_{\widehat K}$ is defined as the minimizer of the penalized cost over the set of all partitions where every segment has length at least $\delta_T$. We assume $\delta_T$ satisfies: $$\delta_T \asymp \sqrt{T \log T} \quad \text{and} \quad \delta_T \le \ell_T / 3.$$ \end{assumption}

\begin{assumption}[Signal dominance]\label{A7prime} Let \(
\lambda_T = 4\sqrt{2}\,M\sqrt{(8m+5)\log T}
\) and
\(\overline B_T=(4m{+}2)M+\frac{(2m^{2}{+}2m)M}{\delta_T}\). There exists $T_0$ such that for all $T \ge T_0$, \[ \frac{\delta_T}{2} \,\Delta_\star^2 \;>\; \beta_T \;+\; 3 \lambda_T \sqrt{T} \;+\; \overline B_T. \] \end{assumption}

\begin{assumption}[Detectability on mixed intervals]\label{A7}
There exist constants $c_0>0$, $C_m\ge0$, and $T_0$ such that for all $T\ge T_0$,
for every pair of consecutive change points $(\tau_k,\tau_{k+1})$ and interval
$[s,e]$ with $s\le \tau_k<\tau_{k+1}\le e$ and $e - s + 1 \;\ge\; 2\,\delta_T$,
\[
\max_{t \in \mathcal T_{k,s,e}}
\Bigl\{C(s,e)-C(s,t)-C(t+1,e)\Bigr\}
\;\ge\]
\[c_0\,g_k\,\Delta_\star^2 - C_m,
\quad g_k:=\tau_{k+1}-\tau_k.
\]

where
$\mathcal T_{k,s,e}
:=
\bigl\{ t \in [\tau_k,\tau_{k+1}-1]
:\ t - s + 1 \ge \delta_T,\; e - t \ge \delta_T \bigr\}
$
is the set of admissible split points inside $[\tau_k,\tau_{k+1}]$
for which both subsegments $[s,t]$ and $[t+1,e]$ have length at least $\delta_T$.
\end{assumption}

Assumptions~\ref{A1}–\ref{A5} are standard in kernel change-point analysis. 
Assumption~\ref{A1} allows short-range temporal dependence and assumes stationarity within each block, which is a common regularity condition. 
Assumption~\ref{A2} (bounded, characteristic kernel) is textbook in MMD/RKHS theory and ensures that the cost is well behaved and that any distributional shift is in principle detectable. 
Assumption~\ref{A3} is a separation condition that enforces a nontrivial gap between consecutive blocks so that changes are identifiable. 
Assumption~\ref{A4} guarantees that each block is long enough for reliable estimation, with a mild rate chosen to simplify uniform concentration under dependence. 
Assumption~\ref{A5} calibrates the penalty at the level of stochastic fluctuations of the empirical cost, preventing severe oversegmentation.

Assumptions~\ref{A6}, \ref{A7prime} and~\ref{A7} are stronger and are only used for the structural and localization results. 
Assumption~\ref{A6} excludes very short segments by enforcing a minimum length at the same order as the concentration rate, which matches the statistical resolution of the problem. 
Assumption~\ref{A7prime} requires that, at that scale, the cumulative jump signal dominates both the penalty and random fluctuations. 
Assumption~\ref{A7} is a detectability condition on mixed intervals that straddle a true change point, ensuring that the best split yields a clear population improvement whenever a genuine change is present. It prevents cancellations so the one-split fit dominates stochastic noise and the penalty, matching the population gain, up to constants. This is reasonable for stationary blocks with a bounded, characteristic kernel.

\subsection{Theoretical Results}\label{sec:theoretical_results}

The first step is to control how well the empirical cost approximates the
population cost, uniformly over all segments. A Bernstein type bound for
each fixed segment is provided by Proposition~\ref{prop:segmentbound} in the
appendix. A union bound over all segments yields:

\begin{lemma}[uniform deviation over all segments]
\label{lem:uniform}
Let Assumptions~\ref{A1} and~\ref{A2} hold.
Let
$
\mathcal E_T
:=
\Bigl\{
  \forall\,1\le s\le e\le T:\ 
  |\widehat C(s,e)-C(s,e)|
  \le \lambda_T\sqrt{e-s+1}
\Bigr\}.
$
Then, for all integers $T\ge 3$, $\Pr(\mathcal E_T)\ge 1-T^{-1}$.
\end{lemma}
\emph{Informally, the lemma ensures that, with high probability, the empirical cost computed from the data is a good approximation of the corresponding population cost for every segment in the sequence, simultaneously.}

As a direct consequence of this concentration and the penalty choice in Assumption~\ref{A5}, we obtain a simple structural property on truly homogeneous regions.

\begin{proposition}[stability on homogeneous segments]
\label{prop:stability-homog}
Let Assumptions~\ref{A1}, \ref{A2}, and \ref{A5} hold.
Then, with probability at least $1-T^{-1}$, the following holds
simultaneously for every segment $[s,e]$ that does not contain a true
change point (that is, $\tau_{k-1} < s \le e < \tau_k$ for some $k$)
and every split point $t$ with $s\le t<e$:
\[
\widehat C(s,e)
\;<\;
\widehat C(s,t)
\;+\;
\widehat C(t{+}1,e)
\;+\;
\beta_T.
\]
\end{proposition}

\emph{In simple terms, in a region where the distribution does not change, inserting an extra change point does not improve the penalized empirical objective, so the procedure has no incentive to create spurious splits inside stationary blocks.}

For our first main result, we compare the population performance of the estimated segmentation to
that of the best segmentation with the same penalty. This result only
requires Assumption~\ref{A1} and \ref{A2}.

\begin{theorem}[oracle inequality]
\label{prop:oracle-main}
Assume that Assumptions~\ref{A1} and~\ref{A2} hold.
With probability at least $1 - T^{-1}$,
$$\sum_{k=1}^{\widehat K+1}
C(\widehat\tau_{k-1}+1,\widehat\tau_k)
+
\beta_T \widehat K
\;\le\;$$
\begin{equation}\label{eq:oracle-ineq}
\inf_{\boldsymbol\tau_{K'}'}
\Bigl\{
\sum_{k=1}^{K'+1}
C(\tau_{k-1}'+1,\tau_k')
+
\beta_T K'
\Bigr\}
+
2 \lambda_T T.
\end{equation}
\end{theorem}

\emph{This result shows that, in terms of the ideal population criterion, our estimator performs almost as well as the best segmentation that could be chosen with full knowledge of the true block distributions, up to a controlled statistical error term arising from Lemma~\ref{lem:uniform}.}

To understand individual change points, we use the stronger assumptions~\ref{A3}--\ref{A7}. A key structural consequence, proved via
Lemma~\ref{lem:a6gap-to-a6dagger} and Lemma~\ref{lem:no-overfull} in the
appendix, is that the estimator does not merge multiple true changes into a
single segment.
\emph{In simple terms, this means that each estimated segment can hide at most one true change point; the procedure does not lump several true changes together into a single segment.}

Combined with a strict improvement property for mixed segments
(Lemma~\ref{lem:strict_improvement_admissible} in the appendix) and the
uniform deviation event $\mathcal E_T$, this leads to the localization
guarantee.

\begin{figure*}[!t]
     \centering
         \centering
         \includegraphics[width=0.95\textwidth]{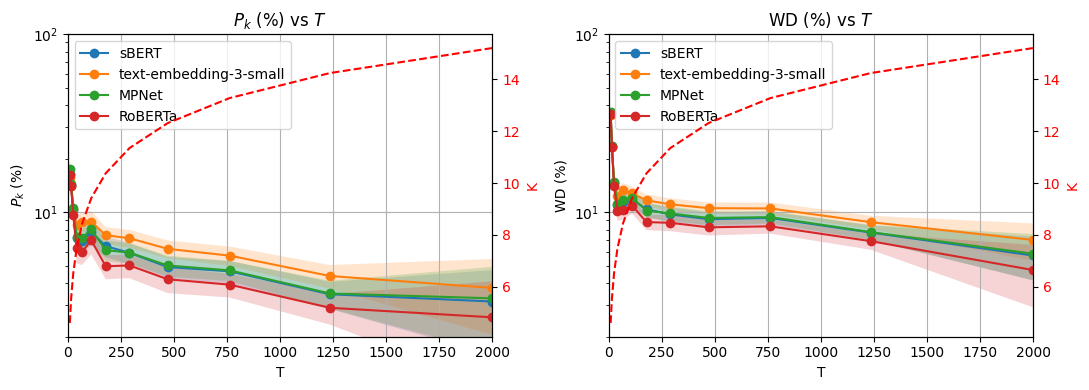}
         \caption{Segmentation accuracies versus sequence length $T$ for Embed-KCPD applied to synthetically generated short-range dependent text data with GPT-4.1 and $m=20$. 
Curves compare three embedding methods (sBERT, MPNet, text-embedding-3-small, RoBERTa). 
Dashed red line shows the growth of the number of change points $K \approx 2\log T$.}\label{fig:mdata}
\vspace{-0.5cm}
 \end{figure*}

\begin{theorem}[localization rate]
\label{thm:localisation-main}
Let Assumptions~\ref{A1}--\ref{A7} hold.
Let $\delta_T$ be the minimum segment length from Assumption~\ref{A6}.
Then as $T \to \infty$,
\begin{equation}
\label{eq:localisation_rate}
\Pr\Bigl(
  \forall\,1 \le k \le K:\ 
  \min_{0 \le j \le \widehat K} |\widehat\tau_j - \tau_k^\star|
  \le \delta_T
\Bigr)
\;\longrightarrow\;
1
\end{equation}
In particular,
\[
\max_{1 \le k \le K} \min_{0 \le j \le \widehat K}
|\widehat\tau_j - \tau_k^\star|
= O_p(\delta_T).
\]
\end{theorem}

This is a particularly relevant consequence of our analysis: \emph{under our signal and spacing assumptions, every true change point is matched by an estimated one within a small window of length $\delta_T$, with probability tending to one. The worst case error is therefore of order $\delta_T$, and since $\delta_T$ is much smaller than the minimal block length $\ell_T$, this means that the error is tiny compared to the size of each stationary segment, so each change point is recovered at an increasingly precise relative position within its block.}

\textbf{Remark.} The $\sqrt{T\log T}$ scaling in $\delta_T$ is a conservative sufficient condition driven by uniform concentration and a single global penalty. Empirically, our Embed-KCPD performs well on datasets with much shorter segments, and the theory should be interpreted as a conservative sanity guarantee under short-range dependence rather than a practical tuning rule.

\section{Embed-KCPD: Instantiation of KCPD for Text Segmentation}\label{sec:text}

We now instantiate Embed-KCPD as a general KCPD framework
for text segmentation. The observed sequence $X_1,\dots,X_T$
consists of contiguous text units (sentences, paragraphs, or dialogue turns).
Each $X_t$ is mapped to a normalized vector representation $
Y_t = f(X_t) \in \mathbb R^d$,
where $f$ is a sentence-embedding model.

In the text setting, change points correspond to topic or discourse
changes that induce distributional shifts in the embedding space.
Assumption~\ref{A1} is natural here:
while consecutive sentences are dependent through syntax and discourse,
dependence decays quickly, and $m$-dependence provides a tractable
abstraction of short-range linguistic correlations.
Assumption~\ref{A3} requires distinct mean embeddings across
segments; this holds whenever topics differ sufficiently in their semantic representation.
Assumption~\ref{A4} enforces a minimum segment length,
excluding degenerate cases where boundaries occur after only a few
sentences; in practice this reflects the fact that coherent topics changes usually span multiple sentences.
Finally, Assumption~\ref{A6} corresponds to
boundaries being marked by sufficiently salient semantic shifts that
cannot be explained by local fluctuations.

We implement two kernels $k(y,y')$: a Gaussian RBF, which satisfies Assumption~\ref{A2}, and cosine similarity. We include cosine to align with standard NLP practice for sentence embeddings, even though it violates Assumption~\ref{A2} (it is non-characteristic).

\textbf{Theory–practice gap.}  Our analysis relies on stylized assumptions that act as a tractable proxy for short-range dependence in sequences of sentence embeddings, rather than a literal model of natural language. Consequently, some conditions are worst-case sufficient and likely loose in typical benchmarks. We view the theory as principled support for Embed-KCPD, while the empirical section evaluates performance with pretrained embeddings and efficient kernels (including cosine) under realistic text distributions.

\begin{table*}[!t]
\centering
\caption{Performance of Baselines and Embed-KCPD in Choi's Dataset. The bolded $P_k$ or WD values denote the best performance for each dataset comparing Embed-KCPD with all baselines. $x_y$ denotes mean $x$ with standard deviation $y$. $*$ marks values reported in original papers.} \label{table:choi} 
\begin{tabular}{lcccccccc} 
\toprule
\multirow{2}{*}{\textbf{Methods}} & \multicolumn{2}{c}{\textbf{3-5}} & \multicolumn{2}{c}{\textbf{6-8}} & \multicolumn{2}{c}{\textbf{9-11}} & \multicolumn{2}{c}{\textbf{3-11}} \\ 
\cmidrule(lr){2-3} \cmidrule(lr){4-5} \cmidrule(lr){6-7} \cmidrule(lr){8-9}
 & $P_k\downarrow$ & WD $\downarrow$ & $P_k\downarrow$ & WD $\downarrow$ & $P_k\downarrow$ & WD $\downarrow$ & $P_k\downarrow$ & WD $\downarrow$ \\ 
\specialrule{0.8pt}{0pt}{0pt}
\multicolumn{9}{c}{Unsupervised Methods} \\ 
\hline
\multicolumn{9}{@{}l}{\textit{Embed-KCPD (sBERT)}} \\
\quad\quad Cosine kernel &$5.2_{5.1}$ & $5.2_{5.1}$ & $3.3_{3.6}$ & $3.4_{3.8}$ & $4.1_{4.6}$& $4.2_{4.7}$& $5.7_{5.3}$ & $5.9_{5.4}$  \\ 
\quad\quad RBF kernel & $5.4_{5.1}$ & $5.4_{5.1}$ & $6.7_{5.4}$ & $6.7_{5.5}$ & $7.6_{6.6}$ & $7.6_{6.6}$ & $9.2_{7.0}$ & $9.5_{7.1}$\\ 

\multicolumn{9}{@{}l}{\textit{Embed-KCPD (MPNet)}} \\
\quad\quad Cosine kernel & $4.1_{5.0}$ & $4.1_{5.0}$ &$3.1_{3.6}$ &$3.2_{3.8}$ &$3.8_{4.4}$ &$3.8_{4.4}$ & $5.7_{5.5}$ & $5.9_{5.7}$ \\ 
\quad\quad RBF kernel & $4.4_{5.1}$ & $4.4_{5.1}$ & $5.1_{5.2}$ & $5.1_{5.2}$ & $6.3_{6.6}$ & $6.3_{6.6}$ & $7.7_{6.1}$ & $8.0_{6.3}$ \\ 

\multicolumn{9}{@{}l}{\textit{Embed-KCPD (text-embedding-3-small)}} \\               
\quad\quad Cosine kernel & $\textbf{3.6}_{4.3}$ & $\textbf{3.6}_{4.3}$ & $\textbf{2.5}_{3.3}$& $\textbf{2.6}_{3.4}$ & $3.1_{4.7}$& $3.1_{4.7}$ & $5.2_{5.4}$& $5.4_{5.5}$\\
\quad\quad RBF kernel & $3.9_{4.4}$ & $3.9_{4.4}$ & $4.6_{5.3}$ & $4.7_{5.4}$ & $5.6_{5.2}$ & $7.3_{6.4}$ & $7.6_{6.5}$ & $5.3_{5.3}$\\

\multicolumn{9}{@{}l}{\textit{Embed-KCPD (RoBERTa)}} \\                      
\quad\quad Cosine kernel & $4.1_{4.8}$ & $4.1_{4.8}$ & $2.9_{3.5}$& $3.1_{3.8}$ & $3.4_{4.3}$& $3.6_{4.4}$ & $5.0_{5.2}$& $5.3_{5.4}$\\
\quad\quad RBF kernel & $4.3_{5.0}$ & $4.3_{5.0}$ & $4.9_{5.0}$ & $5.0_{5.0}$ & $5.7_{5.5}$ & $5.7_{5.5}$ & $8.0_{6.2}$ & $8.3_{6.3}$\\
\hline
\multicolumn{9}{@{}l}{\textit{Baselines}} \\
\quad\quad Coherence
  & $4.4^{*}$ & $6.2^{*}$ & $3.1^{*}$ & $3.3^{*}$ 
  & $\textbf{2.5}^{*}$ & $\textbf{2.6}^{*}$ & $\textbf{4.0}^{*}$ & $\textbf{4.4}^{*}$ \\
\quad\quad GraphSeg 
  & $5.6^{*}$ & $8.7^{*}$ & $7.2^{*}$ & $9.4^{*}$ 
  & $6.6^{*}$ & $9.6^{*}$ & $7.2^{*}$ & $9.0^{*}$ \\
\quad\quad TextTiling
  & $44^{*}$ & -- & $43^{*}$ & -- & $48^{*}$ & -- & $46^{*}$ & -- \\
\quad\quad TextTiling (MPNet)
    & $44.6_{5.6}$ & $86.3_{9.6}$
    & $37.6_{6.4}$ & $76.7_{10.1}$
    & $31.1_{5.4}$ & $70.1_{8.8}$ 
    & $31.7_{6.6}$ & $71.5_{9.6}$ \\
\quad\quad TextTiling (sBERT)
    & $50.0_{3.5}$ & $96.9_{3.6}$
    & $45.3_{5.1}$ & $91.7_{4.6}$
    & $40.3_{4.3}$ & $86.6_{5.4}$ 
    & $41.2_{5.9}$ & $86.8_{6.6}$ \\
\hline
Choi \citep{choi-2000-advances} 
  & $12.0^{*}$ & -- & $9.0^{*}$ & -- & $9.0^{*}$ & -- & $12.0^{*}$ & -- \\ 
\citet{brants-2002} 
  & $7.4^{*}$ & -- & $8.0^{*}$ & -- & $6.8^{*}$ & -- & $19.7^{*}$ & -- \\
\citet{Fragkou-2004-Dynamic} 
  & $5.5^{*}$ & -- & $3.0^{*}$ & -- & $1.3^{*}$ & -- & $7.0^{*}$ & -- \\
\citet{misra-2009} 
  & $23.0^{*}$ & -- & $15.8^{*}$ & -- & $14.4^{*}$ & -- & $16.1^{*}$ & -- \\
\bottomrule
\end{tabular} 
\end{table*}

\subsection{Empirical Evidence of Practical Consistency}\label{sec: empirical_evidence}

To assess the practical reach of theory for Embed-KCPD in text segmentation under controlled conditions with flexible assumptions, we design a simulation with synthetic sequences generated by the large language model GPT-4.1.

We first generate five topic-specific documents (\emph{soccer}, \emph{coffee}, \emph{AI}, \emph{travel}, \emph{dogs}), each with 500 sentences. Within each document, sentences are produced sequentially by prompting GPT\!-\!4.1 to add one sentence at a time, conditioning on the previous \(m \in \{10,20,30\}\) sentences and the document topic; this induces short-range dependence with finite memory and provides clean topic coherence. The resulting process is $m$-order Markov rather than strictly 
$m$-dependent, which is often a more realistic abstraction for text, where dependence decays with distance rather than vanishing exactly. Then, for sequence lengths \(T \le 2000\), we set the number of change points to \(K=\lceil 2\log T\rceil\), randomize change-point locations, and assemble each sequence by concatenating segments drawn from a random selection from the five documents such that consecutive segments have different topics. We generate 100 replicates for each \((T,K)\). See details in Appendix~\ref{app:m-dependent data}. Finally, we estimate change points with Embed-KCPD using four sentence-embedding variants and a penalty of the form \(\beta_T = C\,\sqrt{T\log T}\), for $C\in\{0.001,0.01,0.1,1\}$, matching the theorem’s asymptotic scaling. 




\textbf{Evaluation metrics.} Following previous work, we evaluate text segmentation with two standard metrics: $P_k$ \citep{beeferman1999} and WindowDiff (WD) \citep{pevzner-hearst-2002-critique}. $P_k$ measures the probability that two sentences within a fixed window are incorrectly assigned to the same or different segments, while WD compares the number of predicted and true boundaries in each window, penalizing both false positives and false negatives. Lower scores indicate better performance. By default, the window size for both metrics is set to half the average true segment length. We adopt the same metrics for the  experiments in Sec.~\ref{sec: exp_eval}.


\textbf{Results. }Figures~\ref{fig:sensitivity} and~\ref{fig:msensitivity} in Appendix~\ref{app:sims} summarize results on $P_k$ varying $C$ and $m$. The value \(C=0.1\) yields the best stable asymptotic performance as \(T\) increases, consistent with our theoretical scaling. Although this value is smaller than the conservative lower bound in our assumptions, such under-penalization is common in practice: it increases sensitivity to boundaries while preserving the prescribed asymptotic rate. Results also indicate that the asymptotics are not sensitive to the value of $m$, which is in practice unknown.
Full results for \(C=0.1\) and $m=20$ are shown in Fig.~\ref{fig:mdata}. Empirically, \(P_k\) and WD decrease as \(T\) grows (with \(K\) scaling as above), indicating improved segmentation accuracy consistent with our asymptotic guarantees on change-point recovery; despite the theoretical assumptions being only partially satisfied.

\begin{table*}[!t]
\centering
\caption{Performance of Baselines and Embed-KCPD in Wikipedia, Elements and arXiv Dataset. The bolded $P_k$ or WD values denote values where Embed-KCPD surpassed all unsupervised baselines. The last 3 rows serve only as a reference on supervised methods. 
$x_y$ denotes mean $x$ with standard deviation $y$. $*$ indicates values reported from the original papers.} \label{table:other_data} 
\begin{tabular}{lcccccccc} 
\toprule
\multirow{2}{*}{\textbf{Methods}} & \multicolumn{2}{c}{\textbf{Wiki-300}} & \multicolumn{2}{c}{\textbf{Wiki-50}} &
\multicolumn{2}{c}{\textbf{Elements}} &
\multicolumn{2}{c}{\textbf{arXiv}}
\\
\cmidrule(lr){2-3} \cmidrule(lr){4-5} \cmidrule(lr){6-7} \cmidrule(lr){8-9}
 & $P_k\downarrow$ & WD $\downarrow$ & $P_k\downarrow$ & WD $\downarrow$ & $P_k\downarrow$ & WD $\downarrow$ & $P_k\downarrow$ & WD $\downarrow$ \\ 
\specialrule{0.8pt}{0pt}{0pt}
\multicolumn{9}{c}{Unsupervised Methods} \\ 
\hline
\multicolumn{9}{@{}l}{\textit{Embed-KCPD (sBERT)}} \\ 
\quad\quad Cosine kernel
    & \textbf{33.9}$_{13.0}$ & \textbf{35.2}$_{12.3}$ 
    & $42.4_{15.1}$ & $\textbf{43.8}_{15.3}$ 
    & $\textbf{40.0}_{15.9} $  & $47.5_{15.9}$ 
    & \textbf{7.9}$_{7.2}$ & \textbf{8.2}$_{7.6}$\\
\quad\quad RBF kernel 
    & \textbf{34.2}$_{12.6}$ & $\textbf{37.3}_{13.7}$ 
    & $47.2_{17.4}$ & $51.9_{21.0}$ 
    & \textbf{33.3}$_{15.9}$ & $44.0_{16.6}$ 
    & $\textbf{11.2}_{9.5}$ &  $\textbf{11.8}_{10.1}$\\

\multicolumn{9}{@{}l}{\textit{Embed-KCPD (MPNet)}} \\
\quad\quad Cosine kernel 
    & $\textbf{33.2}_{12.6}$ & \textbf{34.4}$_{11.7}$ 
    & $40.5_{16.2}$ & $\textbf{42.0}_{16.8}$ 
    & $\textbf{41.1}_{16.4}$ & $47.6_{16.0}$ 
    & \textbf{9.1}$_{9.1}$ & \textbf{9.2}$_{9.1}$ \\
\quad\quad RBF kernel 
    &  \textbf{34.0}$_{11.9}$ & $\textbf{35.1}_{11.3}$ 
    & $44.8_{17.3}$ & $49.3_{20.1}$ 
    & $\textbf{32.9}_{16.2}$ & $43.3_{16.2}$ 
    & \textbf{14.7}$_{11.1}$ & \textbf{15.7}$_{12.0}$ \\

\multicolumn{9}{@{}l}{\textit{Embed-KCPD (text-embedding-3-small)}} \\
\quad\quad Cosine kernel 
    & \textbf{32.8}$_{12.8}$ & $\textbf{33.8}_{12.0}$ 
    & $\textbf{38.0}_{14.7}$ & $\textbf{39.8}_{15.8}$ 
    & $44.9_{17.4}$ & $50.3_{16.7}$ 
    & \textbf{9.2}$_{9.8}$ & \textbf{9.3}$_{9.9}$ \\
\quad\quad RBF kernel  
    & $\textbf{33.8}_{12.6}$ & \textbf{34.7}$_{12.0}$ 
    & $43.9_{17.0}$ & $48.5_{20.5}$ 
    & \textbf{32.1}$_{16.3}$ & $43.0_{17.4}$ &
    \textbf{11.3}$_{10.6}$ & \textbf{11.7}$_{11.0}$\\

\multicolumn{9}{@{}l}{\textit{Embed-KCPD (RoBERTa)}} \\
\quad\quad Cosine kernel 
    & \textbf{32.4}$_{12.8}$ & $\textbf{33.7}_{11.9}$ 
    & $39.5_{14.7}$ & $\textbf{41.6}_{15.5}$ 
    & $\textbf{37.8}_{18.7}$ & $45.5_{17.9}$ 
    & \textbf{8.3}$_{7.9}$ & \textbf{8.8}$_{8.6}$ \\
\quad\quad RBF kernel  
    & $\textbf{33.1}_{12.9}$ & $\textbf{34.3}_{12.1}$ 
    & $44.5_{17.2}$ & $49.2_{20.4}$ 
    &\textbf{33.8}$_{16.5}$ & $45.4_{16.8}$ 
    &\textbf{10.7}$_{9.4}$ & \textbf{11.7}$_{10.1}$\\
\hline
\multicolumn{9}{@{}l}{\textit{Baselines}} \\

\quad Coherence &$50.2^*$ & $53.4^*$ & $53.5_{12.3}$ & $71.1_{18.4}$ & $42.4_{18.1}$ & $54.7_{16.6}$  & $43.0_{8.9}$ & $45.4_{9.3}$\\
\quad GraphSeg  &$50.7_{11.4}$ &$54.8_{12.8}$ & $50.2_{17.1}$ & $50.9_{18.7}$ & $52.9_{19.3}$ & $42.3_{16.3}$ 
& $29.0_{11.7}$ & $29.1_{12.1}$\\
\quad TextTiling 
    & $60.3_{9.1}$ & $66.3_{11.2}$ 
    & $47.6_{11.8}$ & $48.3_{11.8}$ 
    & $49.6_{18.3}$ & $50.4_{20.5}$ 
    & $47.9_{9.1}$ & $40.1_{7.7}$ \\
\quad TextTiling (MPNet)
    & $38.1_{12.4}$ & $46.0_{14.3}$
    & $38.9_{14.5}$ & $44.6_{15.4}$
    & $60.8_{19.3}$ & $60.8_{19.3}$
    & $27.1_{7.2}$ & $39.9_{7.9}$ \\
\quad TextTiling (sBERT)
    & $41.1_{14.6}$ & $53.8_{18.9}$
    & $40.7_{13.8}$ & $49.8_{19.0}$
    & $60.8_{19.0}$ & $60.9_{18.9}$
    & $34.8_{8.3}$ & $73.7_{10.5}$ \\
\specialrule{0.8pt}{0pt}{0pt}
\multicolumn{9}{c}{\emph{Supervised Methods}} \\ 

\hline
\emph{NTS} & \emph{34.4}$^*$ & \emph{31.5}$^*$ & -- & -- & -- & -- & -- & --\\
\emph{CATS} &-- &-- & \emph{16.5}$^*$ & -- & \emph{18.4}$^*$ & -- & -- & --\\
\emph{TextSeg} & -- &-- & \emph{18.2}$^*$ & --  &\emph{41.6}$^*$ & --& -- & --\\
\bottomrule
\end{tabular}
\end{table*}

\section{Experimental Evaluation}\label{sec: exp_eval}


\textbf{Datasets}. We evaluate our methods on several widely used datasets for text segmentation. Choi's dataset \citep{choi-2000-advances}, consisting of 700 synthetic documents, serves as the benchmark for segmentation performance. Wiki-300, introduced by \citep{Badjatiya-2018}, contains 300 documents. We also include two smaller datasets: Wiki-50 introduced by \citet{koshorek-etal-2018-text} and Elements \citep{chen-etal-2009-global} about 118 chemical elements. In addition, we construct a new dataset of 20 documents by randomly selecting some recent abstracts from arXiv paper and concatenating them to form one document, to add a clean dataset unknown to all baseline methods. A summary of dataset statistics is presented in Table \ref{ref:summary}. The detailed procedure for constructing the arXiv dataset is in Appendix \ref{ref:arxiv}.

\textbf{Experimental details.} 
For each dataset, we apply Embed-KCPD with both the cosine and RBF kernels, using four modern sentence embeddings for text segmentation: sBERT \citep{reimers-gurevych-2019-sbert}, MPNet \citep{mpnet-2020}, text-embedding-3-small \citep{openai_emb}, and RoBERTa \citep{liu2019roberta}. As unsupervised baselines, we include TextTiling \citep{hearst-1994-multi}, GraphSeg \citep{glavas-etal-2016-unsupervised}, and Coherence \citep{coherence-2024}, and compare their performance with Embed-KCPD across all datasets. 
We also compare with a modern version of TextTiling using modern embeddings, following \citet{saeedabc_llm_text_tiling} tuning configuration for sBERT and MPNet embeddings. For the comparison of Choi's dataset \cite{choi-2000-advances}, we further compare other unsupervised methods, \citep{choi-2000-advances, brants-2002, Fragkou-2004-Dynamic, misra-2009}. See Appendix~\ref{app:implementation} for more implementation details. For the Wikipedia-based datasets, we additionally include supervised approaches reported in prior work: NTS \citep{Badjatiya-2018}, CATS \citep{CATs-2020}, TextSeg \citep{koshorek-etal-2018-text}. We use \(\beta_T = C\,\sqrt{T\log T}\). 

We select a single global $C$ using an unsupervised elbow method, setting $C=0.06$ for the RBF kernel and $C=0.088$ for the cosine kernel across all benchmarks (see Appendix~\ref{app:optimal_c} for details). Figure~\ref{fig:c_pk} in Appendix indicates that performance remains stable across a range of $C$ values.



\begin{figure*}[!t]
    \centering
    \includegraphics[width=0.93\linewidth]{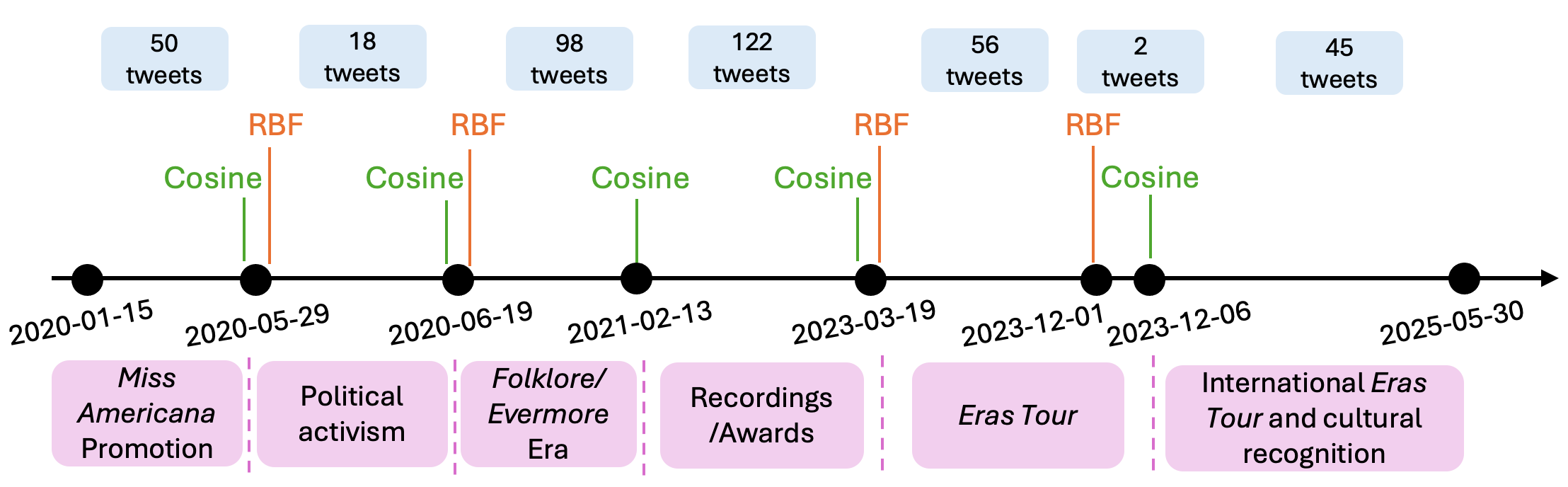}
    \caption{Timeline of Taylor Swift’s tweet stream segmented by Embed-KCPD using RBF and cosine kernels. Each segment is annotated with its tweet count (blue boxes) and an interpretation of its content (pink boxes).}
    \label{fig:timeline_taylor}
        \vspace{-0.5cm}
\end{figure*}

\subsection{Main Results} \label{sec: main_results}

\subsubsection{Results on Choi’s Dataset} Table~\ref{table:choi} reports performance on the synthetic Choi benchmark. Across all settings, Embed-KCPD with a cosine kernel consistently outperforms the RBF kernel, especially for group 3-11, despite the cosine kernel falls outside our theoretical guarantees. This behavior reflects the highly stylized nature of Choi’s dataset: documents are extremely short and segment boundaries are dominated by sharp topic shifts in lexical overlap. In such settings, cosine similarity appears better suited to capturing these discontinuities. 
Among embeddings, text-embedding-3-small combined with cosine kernel achieves the strongest overall performance. 
More generally, Embed-KCPD exhibits stable and consistent performance across kernels and embeddings. While Coherence achieves the best scores on the 3--11 and 9--11 groups, Embed-KCPD delivers competitive results on a dataset that is widely used in the literature, despite being highly artificial relative to real-world text segmentation tasks.

\subsubsection{Results on Other Benchmarks} 
Table~\ref{table:other_data} summarizes results on more realistic datasets: Wiki-300, Wiki-50, Elements, and arXiv. We include supervised methods for reference.

\textbf{Comparing Embed-KCPD to unsupervised baselines}. Embed-KCPD variants outperform all baselines across most datasets and evaluation metrics. As shown in Table \ref{table:other_data}, Embed-KCPD achieves lower $P_k$ and WD in nearly all settings, with few exceptions. 
Importantly, even when TextTiling is augmented with sentence embeddings, Embed-KCPD typically achieves superior performance, indicating that its gains are not solely attributable to the use of embeddings. These results demonstrate the effectiveness of Embed-KCPD as an unsupervised method.

\textbf{Comparing kernels and embeddings.} 
Results with Embed-KCPD using RBF and cosine kernels are more balanced than in Choi’s dataset: the cosine kernel surpasses the RBF on Wiki-300, Wiki-50, and arXiv datasets, while RBF achieves stronger performance on Elements. This variation suggests that our theoretical framework, though developed for characteristic kernels, does not preclude competitive results with alternatives in practice. Among embeddings, text-embedding-3-small yields the lowest $P_k$ and WD on Wiki-50, while RoBERTa achieves the lowest score on the remaining three datasets. Overall, performance differences across embeddings are modest, underscoring the robustness of Embed-KCPD to both kernel and embedding choices.

\textbf{Comparing Embed-KCPD with supervised methods.} 
On Wiki-300, Embed-KCPD achieves lower $P_k$ than \citet{Badjatiya-2018} across all embeddings and kernels, with WD approaching the supervised baseline. On Elements, Embed-KCPD attains lower $P_k$ than \citet{koshorek-etal-2018-text} for most kernel-embedding combinations, with the exception of the kCPD kernel paired with text-embedding-3-small, where performance remains close. These findings suggest that Embed-KCPD, despite being unsupervised, achieves performance comparable to strong supervised methods.

\section{Case Study}\label{sec:case}

To demonstrate Embed-KCPD’s practical value, we include a real-world case study on social-media data: 391 Taylor Swift tweets collected from January 2020 through May 2025. This example shows how a practitioner can readily apply Embed-KCPD to detect topical shifts and conduct downstream analysis in a realistic setting.

\textbf{Experimental details}. 
Consistent with Sec.~\ref{sec: main_results}, we use text-embedding-3-small, which delivers strong segmentation across benchmarks. Following the same procedure used for the benchmark datasets, we choose $C$ via the elbow method (Fig.~\ref{fig:c_taylor} in Appendix), yielding $C=0.03$ for Embed-KCPD with an RBF kernel and 
$C=0.04$ for the cosine kernel. Using these settings, we apply Embed-KCPD to Taylor Swift’s tweet stream with both kernels and analyze the resulting segments. The detected breakpoints appear on the timeline in Fig.~\ref{fig:timeline_taylor}.



\textbf{Interpretation.} 
The first segment aligns with \emph{Miss Americana} promotion and early COVID-19 reflections (Jan–May 2020). The second reflects heightened political engagement (May–Jun 2020). A third segment, captured only by the cosine kernel, covers the \emph{folklore/evermore} era (Jun 2020–Feb 2021), followed by an extended recording/awards period (Feb 2021–Mar 2023). The first year of the famous \emph{Eras Tour} marks the next segment (Mar–Dec 2023). We observe a minor discrepancy between RBF and cosine in the end date of this segment, which we treat as the same change point in practice. The final segment (Dec 2023–May 2025) corresponds to re-releases and broader cultural recognition. Overall, the detected boundaries closely track well-known events in Taylor Swift’s timeline, illustrating Embed-KCPD’s ability to recover meaningful shifts in text streams.

\section{Conclusion}
We performed both a theoretical and empirical study of kernel change-point detection under $m$-dependence by proving an oracle inequality and consistency in change points locations. Building on this, we instantiated Embed-KCPD for unsupervised text segmentation and presented a comprehensive empirical evaluation, demonstrating strong performance against baselines and applicability in a real dataset. In doing so, we bridge theoretical guarantees with practical effectiveness, highlighting Embed-KCPD as an applicable framework for text segmentation.

\section*{Impact Statement}

This paper presents work whose goal is to advance the field of Machine
Learning. There are many potential societal consequences of our work, none
which we feel must be specifically highlighted here.


\bibliography{main}
\bibliographystyle{icml2026}

\newpage
\appendix
\onecolumn

\section{Proofs}\label{app:proof}
\subsection{Auxiliary Results for Lemma 1}

\begin{proposition}[m\textnormal{-dependent concentration for segment cost}]
\label{prop:segmentbound}
Fix integers $1\le s\le e\le T$ and set $n=e-s+1$. Under Assumptions~\ref{A1}–\ref{A2}, for every $x>0$,
\[
\Pr\!\bigl(|\widehat C(s,e)-C(s,e)|>x\bigr)
\;\le\;
4\exp\!\Bigl(-\,\frac{x^{2}}{8(8m+5)M^{2}\,n}\Bigr).
\]
\end{proposition}

\begin{proof}
Write
\[
\widehat C(s,e)-C(s,e)
=\underbrace{\sum_{t=s}^{e}\!\bigl(k(Y_t,Y_t)-\mathbb E[k(Y_t,Y_t)]\bigr)}_{=:A}
-\underbrace{\frac{1}{n}\sum_{i=s}^{e}\sum_{j=s}^{e}\!\bigl(k(Y_i,Y_j)-\mathbb E[k(Y_i,Y_j)]\bigr)}_{=:B}.
\]
Since $0\le k\le M$, each centered summand is bounded in absolute value by $M$.

We use Janson’s inequality for sums with a dependency graph (Thm.~2.1 \citet{janson2004}). 
If $\{X_v\}_{v\in V}$ are centered, $|X_v|\le b$, and $G=(V,E)$ is a dependency graph with chromatic number $\chi(G)$, then for any $t>0$,
\begin{equation}\label{eq:janson}
\Pr\!\Big(\Big|\sum_{v\in V}X_v\Big|>t\Big)\ \le\ 2\exp\!\Big(-\frac{t^2}{2\,\chi(G)\,|V|\,b^2}\Big).
\end{equation}
For $A$, take $V_A=\{s,\dots,e\}$ and connect $t,t'$ when $|t-t'|\le m$. 
This is a valid dependency graph by $m$-dependence (Assumption~\ref{A1}): variables further than $m$ apart are independent. 
The graph is properly colored by $t\bmod (m{+}1)$, hence $\chi(G_A)\le m+1$ and $|V_A|=n$. 
Applying \eqref{eq:janson} with $b=M$ and threshold $t=x/2$ gives
\begin{equation}\label{eq:A}
\Pr\bigl(|A|>x/2\bigr)\ \le\ 2\exp\!\Big(-\frac{x^2}{8(m+1)nM^2}\Big).
\end{equation}
Write $B=\frac{1}{n}S$ with 
\[
S:=\sum_{i=s}^{e}\sum_{j=s}^{e} Z_{ij},\qquad Z_{ij}:=k(Y_i,Y_j)-\mathbb E[k(Y_i,Y_j)].
\]
We consider ordered pairs $(i,j)$ so that $|V_B|=n^2$. 
Define a dependency graph $G_B$ on $V_B=\{(i,j):s\le i,j\le e\}$ by connecting $(i,j)$ and $(i',j')$ iff
\[
\min\{|i-i'|,\,|i-j'|,\,|j-i'|,\,|j-j'|\}\ \le\ m.
\]
Each $Z_{ij}$ is a function of $(Y_i,Y_j)$. If two disjoint vertex sets $U,W\subseteq V_B$ have no edges between them, then the index sets of $Y$’s underlying $U$ and $W$ are pairwise more than $m$ apart in time, hence independent by $m$-dependence; therefore $\{Z_u:u\in U\}$ and $\{Z_w:w\in W\}$ are independent, as required.

Fix $(i,j)$. Let $T_{ij}:=\{k: |k-i|\le m \text{ or } |k-j|\le m\}$; then $|T_{ij}|\le (2m{+}1)+(2m{+}1)=4m+2$.
Any neighbor $(i',j')$ must satisfy $i'\in T_{ij}$ or $j'\in T_{ij}$. Thus the number of neighbors is at most
\[
n\,|T_{ij}|\ +\ n\,|T_{ij}|\ \le\ 2n(4m+2)\ =\ (8m+4)\,n,
\]
so $\Delta(G_B)\le (8m+4)n$ and hence $\chi(G_B)\le \Delta(G_B)+1\le (8m+4)n+1\le (8m+5)n$ for $n\ge 1$.
Applying \eqref{eq:janson} to $S$ with $b=M$, $|V_B|=n^2$, $\chi(G_B)\le (8m+5)n$, and threshold $t=nx/2$ yields
\begin{equation}\label{eq:B}
\Pr\bigl(|B|>x/2\bigr)\ =\ \Pr\bigl(|S|>nx/2\bigr)
\ \le\ 2\exp\!\Big(-\frac{x^2}{8(8m+5)nM^2}\Big).
\end{equation}
If $|\widehat C(s,e)-C(s,e)|=|A-B|>x$, then $|A|>x/2$ or $|B|>x/2$. Hence, by \eqref{eq:A}–\eqref{eq:B},
\[
\Pr\!\bigl(|\widehat C(s,e)-C(s,e)|>x\bigr)
\ \le\ 
2\exp\!\Big(-\frac{x^2}{8(m+1)nM^2}\Big)
+
2\exp\!\Big(-\frac{x^2}{8(8m+5)nM^2}\Big)
\ \le\ 
4\exp\!\Big(-\frac{x^2}{8(8m+5)nM^2}\Big),
\]
where the last inequality uses $(8m+5)\ge (m+1)$ for all $m\ge 0$. This completes the proof.
\end{proof}


\subsection{Proof of Lemma~\ref{lem:uniform} }

Fix $[s,e]$ with length $n=e-s+1$. By Proposition~\ref{prop:segmentbound}, with
$x=\lambda_T\sqrt n$,
\[
\Pr\!\bigl(|\widehat C(s,e)-C(s,e)|>x\bigr)
\ \le\ 4\exp\!\Bigl(-\frac{x^2}{8(8m+5)M^2 n}\Bigr)
\ =\ 4\exp(-4\log T)\ =\ 4T^{-4}.
\]
There are $\frac{T(T+1)}{2}$ segments, so by a union bound,
\[
\Pr(\mathcal E_T^{\mathrm c})
\ \le\ \frac{T(T+1)}{2}\cdot 4T^{-4}
\ =\ \frac{2(T+1)}{T^3}
\ \le\ T^{-1}\qquad\text{for all }T\ge 3,
\]
since $T^2-2T-2\ge 0$ for $T\ge 3$. Hence $\Pr(\mathcal E_T)\ge 1-T^{-1}$.



\subsection{Proof of Proposition~\ref{prop:stability-homog}}

Fix a clean segment $[s,e]$ (i.e., $\tau_{k-1}<s\le e<\tau_k$) and $t\in\{s,\dots,e-1\}$. Write
\[
\Delta\widehat C(a,b):=\widehat C(a,b)-C(a,b),\qquad
\Delta_C:=C(s,t)+C(t{+}1,e)-C(s,e).
\]
We aim to lower bound
\[
\Big[\widehat C(s,t)+\widehat C(t{+}1,e)-\widehat C(s,e)\Big]+\beta_T
=\underbrace{\Delta_C}_{\text{expectation}}+\underbrace{\big(\Delta\widehat C(s,t)+\Delta\widehat C(t{+}1,e)-\Delta\widehat C(s,e)\big)}_{\text{deviation}}+\beta_T.
\]

On the event $\mathcal E_T$ of Lemma~\ref{lem:uniform} (which holds with probability $\ge 1-T^{-1}$), for all $1\le a\le b\le T$,
\[
|\Delta\widehat C(a,b)|\le \lambda_T\sqrt{b-a+1},\qquad
\lambda_T:=4\sqrt{2}\,M\sqrt{(8m+5)\log T}.
\]
Hence, for any $s\le t<e$,
\begin{align*}
\Delta\widehat C(s,t)+\Delta\widehat C(t{+}1,e)-\Delta\widehat C(s,e)
&\ge -\big(|\Delta\widehat C(s,t)|+|\Delta\widehat C(t{+}1,e)|+|\Delta\widehat C(s,e)|\big)\\
&\ge -\lambda_T\!\big(\sqrt{t-s+1}+\sqrt{e-t}+\sqrt{e-s+1}\big)\\
&\ge -3\lambda_T\sqrt{T}.
\end{align*}

Because $[s,e]$ lies within a single stationary block (Assumption~\ref{A1}), $C(s,e)$ depends only on the length $n:=e-s+1$. Denote $C(n):=C(s,e)$.
Set $n_1:=t-s+1$, $n_2:=e-t$, so $n=n_1+n_2$.
For a stationary segment of length $n$,
\begin{equation}\label{eq:Cn}
C(n)
=(n-1)c_0-\frac{2}{n}\sum_{l=1}^{n-1}(n-l)c_l,
\quad\text{where}\quad c_l:=\mathbb E\big[k(Y_1,Y_{1+l})\big].
\end{equation}
Under $m$-dependence, $Y_1$ and $Y_{1+l}$ are independent for $l>m$, hence by bilinearity of the RKHS inner product (no “characteristic” property needed),
\[
c_l=\mathbb E\,\langle \phi(Y_1),\phi(Y_{1+l})\rangle_{\mathcal H}
=\langle \mathbb E\,\phi(Y_1),\mathbb E\,\phi(Y_{1+l})\rangle_{\mathcal H}
=\|\mu_P\|_{\mathcal H}^2=:c_\infty\qquad(l>m).
\]
Define $\delta_l:=c_l-c_\infty$; then $\delta_l=0$ for $l>m$ and, since $|k|\le M$ (Assumption~\ref{A2}), we have $|c_l|\le M$, $|c_\infty|\le M$, thus $|\delta_l|\le 2M$.
Plugging $c_l=c_\infty+\delta_l$ into \eqref{eq:Cn} and using $\sum_{l=1}^{n-1}(n-l)=\tfrac{n(n-1)}{2}$ yields
\[
C(n)=(n-1)(c_0-c_\infty)-2\sum_{l=1}^{\min(n-1,m)}\!\left(1-\frac{l}{n}\right)\delta_l.
\]
Let $V_P:=c_0-c_\infty$ and $S(k):=\sum_{l=1}^{\min(k-1,m)}(1-l/k)\,\delta_l$. Then
\[
\Delta_C=C(n_1)+C(n_2)-C(n_1+n_2)=-V_P-2\big(S(n_1)+S(n_2)-S(n)\big),\quad n=n_1+n_2.
\]
Since $|\delta_l|\le 2M$ and $(1-l/k)\in[0,1]$, we have $|S(k)|\le \sum_{l=1}^{m}|\delta_l|\le 2mM$ for all $k\ge 1$. Also $|V_P|=|c_0-c_\infty|\le 2M$. Therefore
\[
|\Delta_C|\le |V_P|+2\big(|S(n_1)|+|S(n_2)|+|S(n)|\big)\le 2M+2(3\cdot 2mM)=2M(1+6m)=:C_K.
\]

On $\mathcal E_T$,
\[
\big[\widehat C(s,t)+\widehat C(t{+}1,e)-\widehat C(s,e)\big]+\beta_T
\ \ge\ -C_K-3\lambda_T\sqrt{T}+\beta_T.
\]
By Assumption~\ref{A5},
\[
\beta_T\ \ge\ 16M\sqrt{2(8m+5)T\log T}\ +\ 2M(1+6m)\ =\ 4\lambda_T\sqrt{T}+C_K,
\]
so the RHS is at least $\lambda_T\sqrt{T}>0$. Hence
\[
\widehat C(s,e)\ <\ \widehat C(s,t)+\widehat C(t{+}1,e)+\beta_T.
\]

Since $\mathcal E_T$ holds with probability $\ge 1-T^{-1}$ and all bounds above are uniform in $[s,e]$ and $t$, the result holds simultaneously for all clean segments and all splits with that probability.

\subsection{Proof of Theorem~\ref{prop:oracle-main}}

Define the empirical penalized criterion
\[
L(\boldsymbol\tau_{K'}')
:=
\sum_{k=1}^{K'+1}
\widehat C(\tau_{k-1}'+1,\tau_k')
+
\beta_T K'
\]
and the corresponding population penalized criterion
\[
L^\star(\boldsymbol\tau_{K'}')
:=
\sum_{k=1}^{K'+1}
C(\tau_{k-1}'+1,\tau_k')
+
\beta_T K'.
\]
We work on the event \(\mathcal E_T\), which holds with probability at least
\(1 - T^{-1}\).

\paragraph{Step 1: deviation bound for any fixed segmentation.}

Fix an arbitrary segmentation \(\boldsymbol\tau_{K'}'\). For each
\(k \in \{1,\dots,K'+1\}\), let
\[
n_k := \tau_k' - \tau_{k-1}'
\quad\text{so that}\quad
\sum_{k=1}^{K'+1} n_k = T.
\]
The segment \([\tau_{k-1}'+1,\tau_k']\) has length \(n_k\). On \(\mathcal E_T\),
the uniform deviation bound gives
\[
\bigl|\widehat C(\tau_{k-1}'+1,\tau_k') - C(\tau_{k-1}'+1,\tau_k')\bigr|
\le
\lambda_T \sqrt{n_k}
\quad\text{for all }k.
\]

Summing this over all segments, we obtain
\begin{align*}
\biggl|
\sum_{k=1}^{K'+1}
\widehat C(\tau_{k-1}'+1,\tau_k')
-
\sum_{k=1}^{K'+1}
C(\tau_{k-1}'+1,\tau_k')
\biggr|
&\le
\sum_{k=1}^{K'+1}
\bigl|
\widehat C(\tau_{k-1}'+1,\tau_k') - C(\tau_{k-1}'+1,\tau_k')
\bigr|
\\
&\le
\lambda_T \sum_{k=1}^{K'+1} \sqrt{n_k}.
\end{align*}

By the Cauchy--Schwarz inequality,
\[
\sum_{k=1}^{K'+1} \sqrt{n_k}
\le
\sqrt{(K'+1)\sum_{k=1}^{K'+1} n_k}
=
\sqrt{(K'+1)\,T}.
\]
Hence
\[
\biggl|
\sum_{k=1}^{K'+1}
\widehat C(\tau_{k-1}'+1,\tau_k')
-
\sum_{k=1}^{K'+1}
C(\tau_{k-1}'+1,\tau_k')
\biggr|
\le
\lambda_T \sqrt{(K'+1)\,T}.
\]

Since \(K'+1 \le T\) for any segmentation (there can be at most \(T-1\) change
points), we have the simpler bound
\begin{equation}
\label{eq:seg-dev}
\biggl|
\sum_{k=1}^{K'+1}
\widehat C(\tau_{k-1}'+1,\tau_k')
-
\sum_{k=1}^{K'+1}
C(\tau_{k-1}'+1,\tau_k')
\biggr|
\le
\lambda_T T.
\end{equation}

For the penalized criteria this implies
\begin{equation}
\label{eq:L-dev}
\bigl|
L(\boldsymbol\tau_{K'}') - L^\star(\boldsymbol\tau_{K'}')
\bigr|
=
\biggl|
\sum_{k=1}^{K'+1}
\widehat C(\tau_{k-1}'+1,\tau_k')
-
\sum_{k=1}^{K'+1}
C(\tau_{k-1}'+1,\tau_k')
\biggr|
\le
\lambda_T T,
\end{equation}
since the penalty term \(\beta_T K'\) is identical in both \(L\) and
\(L^\star\).

\paragraph{Step 2: comparison between the empirical minimizer and a competitor.}

Let \(\widehat{\boldsymbol\tau}_{\widehat K}\) be any minimizer of \(L\) over all
segmentations. Fix an arbitrary competitor \(\boldsymbol\tau_{K'}'\). We derive
a chain of inequalities on \(\mathcal E_T\).

First, apply \eqref{eq:L-dev} with \(\boldsymbol\tau_{K'}' =
\widehat{\boldsymbol\tau}_{\widehat K}\) to obtain
\begin{equation}
\label{eq:dev-hat}
L^\star(\widehat{\boldsymbol\tau}_{\widehat K})
=
L(\widehat{\boldsymbol\tau}_{\widehat K})
-
\Bigl[
\sum_{k=1}^{\widehat K+1}
\widehat C(\widehat\tau_{k-1}+1,\widehat\tau_k)
-
\sum_{k=1}^{\widehat K+1}
C(\widehat\tau_{k-1}+1,\widehat\tau_k)
\Bigr]
\le
L(\widehat{\boldsymbol\tau}_{\widehat K}) + \lambda_T T.
\end{equation}

Second, by the optimality of \(\widehat{\boldsymbol\tau}_{\widehat K}\) for the
empirical criterion,
\begin{equation}
\label{eq:opt-emp}
L(\widehat{\boldsymbol\tau}_{\widehat K})
\le
L(\boldsymbol\tau_{K'}').
\end{equation}

Third, apply \eqref{eq:L-dev} with \(\boldsymbol\tau_{K'}'\) as given to get
\begin{equation}
\label{eq:dev-comp}
L(\boldsymbol\tau_{K'}')
=
L^\star(\boldsymbol\tau_{K'}')
+
\Bigl[
\sum_{k=1}^{K'+1}
\widehat C(\tau_{k-1}'+1,\tau_k')
-
\sum_{k=1}^{K'+1}
C(\tau_{k-1}'+1,\tau_k')
\Bigr]
\le
L^\star(\boldsymbol\tau_{K'}') + \lambda_T T.
\end{equation}

Combining \eqref{eq:dev-hat}, \eqref{eq:opt-emp}, and \eqref{eq:dev-comp}, we
obtain
\[
\begin{aligned}
L^\star(\widehat{\boldsymbol\tau}_{\widehat K})
&\le
L(\widehat{\boldsymbol\tau}_{\widehat K}) + \lambda_T T
\\
&\le
L(\boldsymbol\tau_{K'}') + \lambda_T T
\\
&\le
L^\star(\boldsymbol\tau_{K'}') + 2 \lambda_T T.
\end{aligned}
\]

Since this holds for an arbitrary competitor \(\boldsymbol\tau_{K'}'\), we can
take the infimum over all segmentations to get
\[
L^\star(\widehat{\boldsymbol\tau}_{\widehat K})
\le
\inf_{\boldsymbol\tau_{K'}'}
L^\star(\boldsymbol\tau_{K'}')
+
2 \lambda_T T.
\]

Unwrapping the definition of \(L^\star\), this inequality is exactly
\eqref{eq:oracle-ineq}:
\[
\sum_{k=1}^{\widehat K+1}
C(\widehat\tau_{k-1}+1,\widehat\tau_k)
+
\beta_T \widehat K
\;\le\;
\inf_{\boldsymbol\tau_{K'}'}
\Bigl\{
\sum_{k=1}^{K'+1}
C(\tau_{k-1}'+1,\tau_k')
+
\beta_T K'
\Bigr\}
+
2 \lambda_T T.
\]

We have proved that the inequality holds on the event \(\mathcal E_T\),
which has probability at least \(1 - T^{-1}\) by Lemma~\ref{lem:uniform}. This completes the proof.

\subsection{Additional Results for Theorem \ref{thm:localisation-main}}

\begin{lemma}[Signal strength on a mixed segment]\label{lem:signal_adjusted}
Let $[s,e]$ contain exactly one true change-point $\tau_k$ with $s \le \tau_k < e$.
Define
\[
n_1:=\tau_k-s+1,\qquad n_2:=e-\tau_k,\qquad n:=n_1+n_2,\qquad \rho:=\frac{n_1 n_2}{n}.
\]
Under Assumptions~\ref{A1}--\ref{A3},
\begin{equation}\label{eq:signal_adjusted_1}
 C(s,e)-C(s,\tau_k)-C(\tau_k{+}1,e)\ \ge\ \rho\,\Delta_k^{2}\;-\;\Bigl((4m{+}2)M+\frac{(2m^{2}{+}2m)M}{n}\Bigr).
\end{equation}
If, in addition, Assumption~\ref{A4} holds and the segment $[s,e]$ satisfies $n_1\ge \ell_T/2$ and $n_2\ge \ell_T/2$, then
\begin{equation}\label{eq:signal_adjusted_2}
 C(s,e)-C(s,\tau_k)-C(\tau_k{+}1,e)\ \ge\ \frac{\Delta_\star^{2}}{4}\,\ell_T\;-\;\Bigl((4m{+}2)M+\frac{(2m^{2}{+}2m)M}{\ell_T}\Bigr).
\end{equation}
\end{lemma}

\begin{proof}
We prove \eqref{eq:signal_adjusted_1} and then deduce \eqref{eq:signal_adjusted_2}.

\paragraph{Part 1: Proof of \eqref{eq:signal_adjusted_1}.}
Using $C(u,v)=\mathbb E[\widehat C(u,v)]$ and expanding the quadratic terms, the diagonal pieces cancel, and we obtain
\[
C(s,e)-C(s,\tau_k)-C(\tau_k{+}1,e)
=
\mathbb E\!\left[
\Bigl(\tfrac{1}{n_1}-\tfrac{1}{n}\Bigr)\!\!\sum_{i,j=s}^{\tau_k}\!k(Y_i,Y_j)
+\Bigl(\tfrac{1}{n_2}-\tfrac{1}{n}\Bigr)\!\!\sum_{i,j=\tau_k+1}^{e}\!k(Y_i,Y_j)
-\tfrac{2}{n}\sum_{i=s}^{\tau_k}\sum_{j=\tau_k+1}^{e} k(Y_i,Y_j)
\right].
\]
Write $\mu_{P_k}:=\mathbb E[\phi(Y)\mid Y\sim P_k]\in\mathcal H$ and recall $k(x,y)=\langle \phi(x),\phi(y)\rangle_{\mathcal H}$.
Introduce the population (independence) proxy by replacing $\mathbb E\,k(Y_i,Y_j)$ with $\langle \mu_{\mathrm{dist}(i)},\mu_{\mathrm{dist}(j)}\rangle_{\mathcal H}$, where $\mathrm{dist}(i)\in\{k,k{+}1\}$ indicates the block of $i$.
This yields the \emph{population term}
\[
\frac{n_1n_2}{n}\,\Bigl(\|\mu_{P_k}\|_{\mathcal H}^2+\|\mu_{P_{k+1}}\|_{\mathcal H}^2-2\langle\mu_{P_k},\mu_{P_{k+1}}\rangle_{\mathcal H}\Bigr)
=\rho\,\|\mu_{P_k}-\mu_{P_{k+1}}\|_{\mathcal H}^{2}
=\rho\,\Delta_k^2,
\]
and a \emph{remainder} $R$ capturing the $m$-dependence corrections.

Let $\delta_{i,j}:=\mathbb E[k(Y_i,Y_j)]-\langle \mu_{\mathrm{dist}(i)},\mu_{\mathrm{dist}(j)}\rangle_{\mathcal H}$. Then $\delta_{i,j}=0$ whenever $|i-j|>m$ by $m$-dependence (Assumption~\ref{A1}); moreover, by boundedness (Assumption~\ref{A2}), $|\delta_{i,j}|\le 2M$.
Writing
\[
R=\frac{n_2}{n n_1}E_1+\frac{n_1}{n n_2}E_2-\frac{2}{n}E_{12},
\]
where $E_1:=\sum_{i,j=s}^{\tau_k}\delta_{i,j}$, $E_2:=\sum_{i,j=\tau_k+1}^{e}\delta_{i,j}$, and $E_{12}:=\sum_{i=s}^{\tau_k}\sum_{j=\tau_k+1}^{e}\delta_{i,j}$, we bound each piece by counting ordered pairs with $|i-j|\le m$:
\begin{align*}
|E_1|&\le \bigl(\text{at most }n_1(2m{+}1)\text{ pairs}\bigr)\cdot 2M
= n_1(2m{+}1)\,2M,\\
|E_2|&\le n_2(2m{+}1)\,2M,\\
|E_{12}|&\le \Bigl(\sum_{d=1}^{m} d\Bigr)\cdot 2M=\frac{m(m{+}1)}{2}\cdot 2M=m(m{+}1)\,M,
\end{align*}
where the last line counts the cross-boundary pairs with offsets $d=1,\dots,m$ once (note the algebra above contributes $-\tfrac{2}{n}E_{12}$, so only left-to-right ordered pairs appear).
Consequently,
\begin{align*}
|R|
&\le \frac{n_2}{n n_1}\,n_1(2m{+}1)\,2M
+\frac{n_1}{n n_2}\,n_2(2m{+}1)\,2M
+\frac{2}{n}\,m(m{+}1)\,M\\
&= \frac{n_1+n_2}{n}\,(4m{+}2)M+\frac{2m^2+2m}{n}M
= (4m{+}2)M+\frac{(2m^2{+}2m)M}{n}.
\end{align*}
Since $C(s,e)-C(s,\tau_k)-C(\tau_k{+}1,e)=\rho\,\Delta_k^2+R$, we obtain \eqref{eq:signal_adjusted_1} from $R\ge -|R|$.

\paragraph{Part 2: Proof of \eqref{eq:signal_adjusted_2}.}
By Assumption~\ref{A3}, $\Delta_k^2\ge \Delta_\star^2$. Under $n_1,n_2\ge \ell_T/2$, the function $\rho=\frac{n_1n_2}{n_1+n_2}$ is minimized at $n_1=n_2=\ell_T/2$, hence
\[
\rho\ \ge\ \frac{(\ell_T/2)^2}{\ell_T}\ =\ \frac{\ell_T}{4}.
\]
Also $n=n_1+n_2\ge \ell_T$, so
\[
-\Bigl((4m{+}2)M+\frac{(2m^2{+}2m)M}{n}\Bigr)
\ \ge\
-\Bigl((4m{+}2)M+\frac{(2m^2{+}2m)M}{\ell_T}\Bigr).
\]
Combining with \eqref{eq:signal_adjusted_1} yields \eqref{eq:signal_adjusted_2}.
\end{proof}

\begin{lemma}[Detectability]\label{lem:a6gap-to-a6dagger}
Let Assumptions~\ref{A1}--\ref{A7} hold and fix $\delta>0$. Then there exists $T_\delta$ such that for all $T\ge T_\delta$ and all intervals
$[s,e]$ containing two consecutive changes $\tau_k<\tau_{k+1}$, there exists
$t^\star\in[\tau_k,\tau_{k+1}-1]$

and $t^\star - s + 1 \ge \delta_T,\; e - t^\star \ge \delta_T $

with
\[
C(s,e)-C(s,t^\star)-C(t^\star+1,e)\ \ge\ \beta_T\ +\ 4\,\lambda_T\,\sqrt{T}\ +\ \delta\,\beta_T,
\]
where $\lambda_T:=4\sqrt{2}\,M\sqrt{(8m+5)\log T}$.
\end{lemma}

\begin{proof}
Fix $\delta>0$. By Assumption~\ref{A7}, for any $[s,e]$ with $s\le \tau_k<\tau_{k+1}\le e$
 there exists
$t^\star\in[\tau_k,\tau_{k+1}-1]$

and $t^\star - s + 1 \ge \delta_T,\; e - t^\star \ge \delta_T $ such that
\begin{equation}\label{eq:gap-lb}
C(s,e)-C(s,t^\star)-C(t^\star+1,e)\ \ge\ c_0\,g_k\,\Delta_\star^2\ -\ C_m .
\end{equation}
By Assumption~\ref{A4}, $g_k\ge \ell_T$, hence
\begin{equation}\label{eq:ell-lb}
c_0\,g_k\,\Delta_\star^2\ -\ C_m\ \ge\ c_0\,\ell_T\,\Delta_\star^2\ -\ C_m .
\end{equation}

From Assumption~\ref{A5} there exists $K_\beta>0$ and $T_1$ such that, for all $T\ge T_1$,
\begin{equation}\label{eq:beta-upper}
\beta_T \ \le\ K_\beta\,\sqrt{T\log T}.
\end{equation}
Moreover, by the definition of $\lambda_T$,
\begin{equation}\label{eq:lambda-upper}
4\,\lambda_T\,\sqrt{T}\ =\ 16\sqrt{2}\,M\,\sqrt{(8m{+}5)\,T\log T}\ =:\ K_2\,\sqrt{T\log T},
\end{equation}
with $K_2:=16\sqrt{2}\,M\sqrt{8m{+}5}$. Therefore, for all $T\ge T_1$,
\begin{equation}\label{eq:target-upper}
(1+\delta)\beta_T + 4\,\lambda_T\sqrt{T}\ \le\ \bigl((1+\delta)K_\beta+K_2\bigr)\,\sqrt{T\log T}.
\end{equation}

Since $\ell_T/\sqrt{T\log T}\to\infty$ by Assumption~\ref{A4}, there exists $T_2$ such that, for all $T\ge T_2$,
\begin{equation}\label{eq:dominance}
c_0\,\ell_T\,\Delta_\star^2\ -\ C_m\ \ge\ \bigl((1+\delta)K_\beta+K_2\bigr)\,\sqrt{T\log T}.
\end{equation}
Combining \eqref{eq:ell-lb}, \eqref{eq:target-upper}, and \eqref{eq:dominance}, we obtain that
for all $T\ge T_\delta:=\max\{T_0,T_1,T_2,3\}$,
\[
c_0\,\ell_T\,\Delta_\star^2\ -\ C_m\ \ge\ (1+\delta)\beta_T + 4\,\lambda_T\sqrt{T}.
\]
Plugging this into \eqref{eq:gap-lb} for the same $t^\star$ yields
\[
C(s,e)-C(s,t^\star)-C(t^\star+1,e)\ \ge\ (1+\delta)\beta_T + 4\,\lambda_T\sqrt{T}
\ =\ \beta_T\ +\ 4\,\lambda_T\sqrt{T}\ +\ \delta\,\beta_T .
\]
All constants are uniform in $k$ and in $[s,e]$ because $\ell_T$, $K_\beta$, and $K_2$ do not depend on $k,[s,e]$.
\end{proof}


\begin{lemma}[No overfull estimated segments]\label{lem:no-overfull}
Let Assumptions~\ref{A1}--\ref{A7} hold. 
Then, with probability at least $1-T^{-1}$, no estimated segment of an optimal
penalized partition contains two true changepoints.
\end{lemma}

\begin{proof}
Let $\widehat{\boldsymbol\tau}_{\widehat K}$ be any minimizer of
\[
L(\boldsymbol{\tau}'_{K'})=\sum_{r=1}^{K'+1}\widehat C(\tau'_{r-1}{+}1,\tau'_r)+\beta_T K'.
\]
Work on the high-probability event
\[
\mathcal E_T:=\Bigl\{\ \forall\,1\le s\le e\le T:\ |\widehat C(s,e)-C(s,e)|\le \lambda_T\sqrt{e-s+1}\ \Bigr\},
\]
which satisfies $\Pr(\mathcal E_T)\ge 1-T^{-1}$ by Lemma~\ref{lem:uniform}.

Suppose, towards a contradiction, that some estimated segment $[s,e]$ induced by
$\widehat{\boldsymbol\tau}_{\widehat K}$ contains two consecutive true
changepoints $\tau_k<\tau_{k+1}$ with $s\le \tau_k<\tau_{k+1}\le e$. 
Fix any $\delta>0$. By Lemma~\ref{lem:a6gap-to-a6dagger}   there exists
$t^\star\in[\tau_k,\tau_{k+1}-1]$

and $t^\star - s + 1 \ge \delta_T,\; e - t^\star \ge \delta_T $ such that
\[
G:=C(s,e)-C(s,t^\star)-C(t^\star{+}1,e)\ \ge\ \beta_T+4\lambda_T\sqrt{T}+\delta\,\beta_T.
\]
On $\mathcal E_T$ we have
\[
\begin{aligned}
\widehat G
&:=\widehat C(s,e)-\widehat C(s,t^\star)-\widehat C(t^\star{+}1,e)\\
&\ge G-\bigl|\widehat C(s,e)-C(s,e)\bigr|
           -\bigl|\widehat C(s,t^\star)-C(s,t^\star)\bigr|
           -\bigl|\widehat C(t^\star{+}1,e)-C(t^\star{+}1,e)\bigr|\\
&\ge G-\lambda_T\!\left(\sqrt{e-s+1}+\sqrt{t^\star-s+1}+\sqrt{e-t^\star}\right).
\end{aligned}
\]
Since $\sqrt{e-s+1}\le \sqrt{t^\star-s+1}+\sqrt{e-t^\star}$ and each square-root term is at most $\sqrt{T}$, we get
\[
\widehat G\ \ge\ G-2\lambda_T\!\left(\sqrt{t^\star-s+1}+\sqrt{e-t^\star}\right)\ \ge\ G-4\lambda_T\sqrt{T}.
\]
Hence, by the lower bound on $G$,
\[
\widehat G\ \ge\ \beta_T+\delta\,\beta_T.
\]
If we refine the partition by inserting a split at $t^\star$,
the data-fit part decreases by $\widehat G$ while the penalty increases by
$\beta_T$, so the net change is
\[
\Delta L\;=\;-\widehat G+\beta_T\ \le\ -(\beta_T+\delta\beta_T)+\beta_T\ =\ -\delta\,\beta_T\ <\ 0,
\]
contradicting optimality of $\widehat{\boldsymbol\tau}_{\widehat K}$. Therefore,
on $\mathcal E_T$, no estimated segment contains two true changepoints. Since
$\Pr(\mathcal E_T)\ge 1-T^{-1}$, the claim follows.
\end{proof}

\begin{corollary}[No estimated segment contains $\ge 2$ true changes]
Let Assumptions~\ref{A1}--\ref{A7} hold. With probability at least $1-T^{-1}$, every estimated segment of an optimal penalized partition contains at most one true changepoint.
\end{corollary}

\begin{proof}
If an estimated segment contained $\ge 2$ true changepoints, it would contain some adjacent pair $(\tau_k,\tau_{k+1})$. Apply Lemma~\ref{lem:a6gap-to-a6dagger} within that segment to obtain a split $t^\star$ such that inserting $t^\star$ strictly decreases the penalized cost, exactly as in the proof of Lemma~\ref{lem:no-overfull}. This contradicts optimality. The high-probability event is the same as in Lemma~\ref{lem:no-overfull}.
\end{proof}

\begin{lemma}[Strict improvement of a mixed segment] \label{lem:strict_improvement_admissible} Let Assumptions~\ref{A1}--\ref{A3}, \ref{A6} and~\ref{A7prime} hold. Let $\mathcal E_T$ be the high probability event from Lemma~\ref{lem:uniform}, \[ \mathcal E_T := \Bigl\{ \forall\,1 \le s \le e \le T: \ |\widehat C(s,e) - C(s,e)| \le \lambda_T \sqrt{e-s+1} \Bigr\}. \] Consider an admissible partition $\boldsymbol\tau$ (that is, all its segments have length at least $\delta_T$) and suppose it contains a segment \[ E = [s,e] = [t_L{+}1, t_R] \] that contains exactly one true change point $\tau_k^\star$ with $s \le \tau_k^\star < e$. Define \[ n_1 := \tau_k^\star - s + 1,\qquad n_2 := e - \tau_k^\star,\qquad n := n_1 + n_2. \] Assume further that \[ n_1 \ge \delta_T \quad\text{and}\quad n_2 \ge \delta_T, \] so that splitting $E$ at $\tau_k^\star$ yields two segments that are still admissible. Then, on the event $\mathcal E_T$, there exists an admissible partition $\boldsymbol\tau'$ (obtained by inserting $\tau_k^\star$ into $\boldsymbol\tau$) such that \[ L(\boldsymbol\tau') < L(\boldsymbol\tau) \] for all sufficiently large $T$. \end{lemma} \begin{proof} We work on the event $\mathcal E_T$ throughout. 

Let $\boldsymbol\tau$ be an admissible partition that has a mixed segment $E=[s,e]$ as in the statement, with $n_1,n_2 \ge \delta_T$. Define a new partition $\boldsymbol\tau'$ by inserting the true change point $\tau_k^\star$ into $\boldsymbol\tau$ inside the segment $E$. That is, we replace the single segment $[s,e]$ by the two segments $[s,\tau_k^\star]$ and $[\tau_k^\star{+}1,e]$, leaving all other segments unchanged. Because \[ n_1 = \tau_k^\star - s + 1 \ge \delta_T \quad\text{and}\quad n_2 = e - \tau_k^\star \ge \delta_T, \] every segment in $\boldsymbol\tau'$ still has length at least $\delta_T$. Thus $\boldsymbol\tau'$ is admissible under Assumption~\ref{A6}. If $\boldsymbol\tau$ has $K$ change points, then $\boldsymbol\tau'$ has $K+1$ change points, so the penalty term in $L$ increases by $\beta_T$. 

By definition of $L$, the only changes come from the segment $E$ and the penalty term. Denote \[ \Delta L := L(\boldsymbol\tau') - L(\boldsymbol\tau). \] We have \[ \Delta L = \beta_T + \widehat C(s,\tau_k^\star) + \widehat C(\tau_k^\star{+}1,e) - \widehat C(s,e). \] Introduce the population change \[ \Delta C_{\mathrm{pop}} := C(s,\tau_k^\star) + C(\tau_k^\star{+}1,e) - C(s,e), \] and the empirical fluctuation \[ \Delta_{\mathrm{noise}} := \bigl(\widehat C(s,\tau_k^\star) - C(s,\tau_k^\star)\bigr) + \bigl(\widehat C(\tau_k^\star{+}1,e) - C(\tau_k^\star{+}1,e)\bigr) - \bigl(\widehat C(s,e) - C(s,e)\bigr). \] Then \[ \Delta L = \beta_T + \Delta C_{\mathrm{pop}} + \Delta_{\mathrm{noise}}. \] We now control $\Delta C_{\mathrm{pop}}$ and $\Delta_{\mathrm{noise}}$. 

The segment $[s,e]$ contains exactly one true change point $\tau_k^\star$, so Lemma~\ref{lem:signal_adjusted}, inequality~\eqref{eq:signal_adjusted_1}, applies: \[ C(s,e) - C(s,\tau_k^\star) - C(\tau_k^\star{+}1,e) \;\ge\; \rho\,\Delta_k^{2} \;-\; \Bigl((4m{+}2)M+\frac{(2m^{2}{+}2m)M}{n}\Bigr), \] where \[ \rho := \frac{n_1 n_2}{n}, \quad n = e-s+1, \quad n_1 = \tau_k^\star - s + 1, \quad n_2 = e - \tau_k^\star. \] Rewriting, we obtain \[ \Delta C_{\mathrm{pop}} = C(s,\tau_k^\star) + C(\tau_k^\star{+}1,e) - C(s,e) \le -\rho\,\Delta_k^2 + \Bigl((4m{+}2)M+\frac{(2m^{2}{+}2m)M}{n}\Bigr). \] By Assumption~\ref{A3}, $\Delta_k^2 \ge \Delta_\star^2$, so \[ \Delta C_{\mathrm{pop}} \le -\rho\,\Delta_\star^2 + \Bigl((4m{+}2)M+\frac{(2m^{2}{+}2m)M}{n}\Bigr). \] Without loss of generality suppose $n_1 \le n_2$. Then \[ \rho = \frac{n_1 n_2}{n_1+n_2} = n_1 \cdot \frac{n_2}{n_1+n_2}. \] Since $n_2 \ge n_1$, we have \[ \frac{n_2}{n_1+n_2} \ge \frac{1}{2}, \] hence \[ \rho \ge \frac{n_1}{2} = \frac{\min(n_1,n_2)}{2}. \] Using $n_1,n_2 \ge \delta_T$, we obtain \[ \rho \ge \frac{\delta_T}{2}. \] Consequently, \[ \Delta C_{\mathrm{pop}} \le - \frac{\delta_T}{2}\,\Delta_\star^2 + \Bigl((4m{+}2)M+\frac{(2m^{2}{+}2m)M}{n}\Bigr). \] Since $E$ is admissible, $n = e-s+1 \ge \delta_T$. Hence \[ (4m{+}2)M+\frac{(2m^{2}{+}2m)M}{n} \;\le\; (4m{+}2)M+\frac{(2m^{2}{+}2m)M}{\delta_T} = \overline B_T. \] We thus obtain \[ \Delta C_{\mathrm{pop}} \le - \frac{\delta_T}{2}\,\Delta_\star^2 + \overline B_T. \] 

On the event $\mathcal E_T$, Lemma~\ref{lem:uniform} gives, for any segment $[u,v]$ of length $n_{uv}=v-u+1$, \[ |\widehat C(u,v) - C(u,v)| \le \lambda_T \sqrt{n_{uv}}. \] Apply this to the three segments: \[ [s,\tau_k^\star] \text{ of length } n_1,\qquad [\tau_k^\star{+}1,e] \text{ of length } n_2,\qquad [s,e] \text{ of length } n = n_1+n_2. \] We obtain \[ \bigl|\widehat C(s,\tau_k^\star) - C(s,\tau_k^\star)\bigr| \le \lambda_T \sqrt{n_1}, \] \[ \bigl|\widehat C(\tau_k^\star{+}1,e) - C(\tau_k^\star{+}1,e)\bigr| \le \lambda_T \sqrt{n_2}, \] \[ \bigl|\widehat C(s,e) - C(s,e)\bigr| \le \lambda_T \sqrt{n}. \] Therefore \[ |\Delta_{\mathrm{noise}}| \le \lambda_T\sqrt{n_1} + \lambda_T\sqrt{n_2} + \lambda_T\sqrt{n} = \lambda_T\bigl(\sqrt{n_1} + \sqrt{n_2} + \sqrt{n}\bigr). \] Using Cauchy--Schwarz, \[ \sqrt{n_1} + \sqrt{n_2} \le \sqrt{2(n_1+n_2)} = \sqrt{2n}, \] so \[ \sqrt{n_1} + \sqrt{n_2} + \sqrt{n} \le \sqrt{2n} + \sqrt{n} \le (\sqrt{2}+1)\sqrt{n} < 3\sqrt{n}. \] Thus \[ |\Delta_{\mathrm{noise}}| \le 3 \lambda_T \sqrt{n} \le 3 \lambda_T \sqrt{T}, \] since $n \le T$. 

Combining the previous bounds, we have \[ \Delta L = \beta_T + \Delta C_{\mathrm{pop}} + \Delta_{\mathrm{noise}} \le \beta_T + \Bigl(- \frac{\delta_T}{2}\,\Delta_\star^2 + \overline B_T\Bigr) + |\Delta_{\mathrm{noise}}|. \] Using the bound on the noise term, \[ \Delta L \le \beta_T - \frac{\delta_T}{2}\,\Delta_\star^2 + \overline B_T + 3 \lambda_T \sqrt{T}. \] By Assumption~\ref{A7prime}, there exists $T_0$ such that for all $T \ge T_0$, \[ \frac{\delta_T}{2}\,\Delta_\star^2 \;>\; \beta_T + \overline B_T + 3 \lambda_T \sqrt{T}. \] Fix any $T \ge T_0$ and define \[ \eta_T := \frac{\delta_T}{2}\,\Delta_\star^2 \;-\; \bigl(\beta_T + \overline B_T + 3 \lambda_T \sqrt{T}\bigr) \;>\; 0. \] Then \[ \beta_T - \frac{\delta_T}{2}\,\Delta_\star^2 + \overline B_T + 3 \lambda_T \sqrt{T} = -\eta_T < 0, \] and hence, using the previous bound on $\Delta L$, \[ \Delta L \le \beta_T - \frac{\delta_T}{2}\,\Delta_\star^2 + \overline B_T + 3 \lambda_T \sqrt{T} = -\eta_T < 0. \] Therefore $L(\boldsymbol\tau') < L(\boldsymbol\tau)$ for all $T \ge T_0$ on the event $\mathcal E_T$. Therefore, for all sufficiently large $T$ and on $\mathcal E_T$, \[ \Delta L \le -\eta_T < 0. \] Hence $L(\boldsymbol\tau') < L(\boldsymbol\tau)$, which shows that the admissible partition obtained by inserting the true change point $\tau_k^\star$ into the mixed segment $E$ has strictly smaller penalized cost than $\boldsymbol\tau$. \end{proof}

\subsection{Proof of Theorem~\ref{thm:localisation-main}}\label{app:prooflocalization}

Recall the high-probability event $\mathcal E_T$ from Lemma~\ref{lem:uniform}:
\[
\mathcal E_T
:=
\Bigl\{
  \forall\,1 \le s \le e \le T:\ 
  |\widehat C(s,e) - C(s,e)|
  \le \lambda_T \sqrt{e-s+1}
\Bigr\},
\]
where $\lambda_T = 4\sqrt{2}\,M\sqrt{(8m+5)\log T}$, and $\Pr(\mathcal E_T) \ge 1 - T^{-1}$ for all $T \ge 3$.

Let $\mathcal N_T$ be the high-probability event from Lemma~\ref{lem:no-overfull} (No overfull estimated segments). That lemma states that, under Assumptions~\ref{A1}--\ref{A7}, for all $T$ large enough
\[
\Pr(\mathcal N_T) \ge 1 - T^{-1},
\]
and on $\mathcal N_T$, \emph{no segment of an optimal penalised partition contains two true change points}.

Define
\[
\Omega_T := \mathcal E_T \cap \mathcal N_T.
\]
Then $\Pr(\Omega_T) \ge 1 - 2 T^{-1} \to 1$ as $T \to \infty$. We will show that on $\Omega_T$ and for $T$ large enough,
\begin{equation}\label{eq:local_goal}
\forall\,1 \le k \le K:\ 
\min_{0 \le j \le \widehat K} |\widehat\tau_j - \tau_k^\star|
\le \delta_T.
\end{equation}
This will imply \eqref{eq:localisation_rate}, because $\Pr(\Omega_T) \to 1$.

So fix $T$ large and suppose $\Omega_T$ holds. Let $\widehat{\boldsymbol\tau}_{\widehat K} \in \mathcal P_T(\delta_T)$ denote the optimal penalized partition (the Embed-KCPD estimator).

We now argue by contradiction. Suppose that there exists at least one true change point that is not localized within $\delta_T$. That is, assume there exists
\[
k^\star \in \{1,\dots,K\}
\quad\text{such that}\quad
\min_{0 \le j \le \widehat K} |\widehat\tau_j - \tau_{k^\star}^\star|
> \delta_T.
\]
Fix such an index $k^\star$.

Let $E$ be the segment of the estimated partition $\widehat{\boldsymbol\tau}_{\widehat K}$ that contains $\tau_{k^\star}^\star$. Concretely, there exists $r \in \{1,\dots,\widehat K+1\}$ such that
\[
E = [s,e] = [\widehat\tau_{r-1}+1,\widehat\tau_r],
\]
with the convention $\widehat\tau_0 = 0$, $\widehat\tau_{\widehat K+1} = T$, and
\[
\widehat\tau_{r-1} < \tau_{k^\star}^\star \le \widehat\tau_r.
\]

By Lemma~\ref{lem:no-overfull}, on $\mathcal N_T$ no optimal penalized segment contains two true change points. Since $E$ contains $\tau_{k^\star}^\star$, it must therefore contain \emph{exactly one} true change point, namely $\tau_{k^\star}^\star$. Thus $E$ is a \emph{mixed} segment with exactly one true change.

Next, because $\tau_{k^\star}^\star$ is at distance strictly greater than $\delta_T$ from every estimated change point, we have
\begin{align*}
\tau_{k^\star}^\star - \widehat\tau_{r-1}
&> \delta_T,
\\
\widehat\tau_r - \tau_{k^\star}^\star
&> \delta_T.
\end{align*}
In terms of subsegment lengths inside $E$, define
\[
n_1 := \tau_{k^\star}^\star - s + 1
= \tau_{k^\star}^\star - \widehat\tau_{r-1},
\qquad
n_2 := e - \tau_{k^\star}^\star
= \widehat\tau_r - \tau_{k^\star}^\star.
\]
Then
\begin{equation}\label{eq:n1n2_large}
n_1 \ge \delta_T + 1 > \delta_T,
\qquad
n_2 \ge \delta_T + 1 > \delta_T.
\end{equation}
Since $\delta_T \to \infty$, for $T$ large enough we can simply write $n_1 \ge \delta_T$ and $n_2 \ge \delta_T$.

Note also that $E$ itself is admissible by construction, since $\widehat{\boldsymbol\tau}_{\widehat K} \in \mathcal P_T(\delta_T)$ implies that $e-s+1 = \widehat\tau_r - \widehat\tau_{r-1} \ge \delta_T$.

Because $E$ contains exactly one true change point $\tau_{k^\star}^\star$ and $n_1, n_2 \ge \delta_T$, and both $E$ and the partition are admissible, we are exactly in the setting of Lemma~\ref{lem:strict_improvement_admissible} (Strict improvement of a mixed segment, admissible split). More precisely, Lemma~\ref{lem:strict_improvement_admissible} applies to:

- the partition $\boldsymbol\tau := \widehat{\boldsymbol\tau}_{\widehat K}$, which belongs to $\mathcal P_T(\delta_T)$ by Assumption~\ref{A6};
- the segment $E = [s,e]$, which is an element of that partition and contains exactly one true change $\tau_{k^\star}^\star$;
- the subsegment lengths $n_1, n_2$ which satisfy $n_1 \ge \delta_T$ and $n_2 \ge \delta_T$.

Lemma~\ref{lem:strict_improvement_admissible} states: on the event $\mathcal E_T$ and under Assumptions~\ref{A1}--\ref{A3}, \ref{A6} and~\ref{A7prime}, for all sufficiently large $T$, there exists a new admissible partition $\boldsymbol\tau'$ obtained from $\boldsymbol\tau$ by \emph{inserting} the true change point $\tau_{k^\star}^\star$ into the segment $E$ such that
\[
L(\boldsymbol\tau') < L(\boldsymbol\tau).
\]

In particular, since we are on $\Omega_T \subseteq \mathcal E_T$, and $T$ is large, when we take $\boldsymbol\tau = \widehat{\boldsymbol\tau}_{\widehat K}$, Lemma~\ref{lem:strict_improvement_admissible} yields an admissible partition $\boldsymbol\tau'$ with
\[
L(\boldsymbol\tau') < L(\widehat{\boldsymbol\tau}_{\widehat K}).
\]

But this contradicts the definition of $\widehat{\boldsymbol\tau}_{\widehat K}$ as an optimal minimizer of $L(\cdot)$ over $\mathcal P_T(\delta_T)$.

Therefore, our assumption that there exists $k^\star$ with
\[
\min_{0 \le j \le \widehat K} |\widehat\tau_j - \tau_{k^\star}^\star| > \delta_T
\]
must be false on $\Omega_T$ for all sufficiently large $T$.

Equivalently, on $\Omega_T$ and for all large $T$,
\[
\forall\,1 \le k \le K:\ 
\min_{0 \le j \le \widehat K} |\widehat\tau_j - \tau_k^\star|
\le \delta_T.
\]
Since $\Pr(\Omega_T) \to 1$ as $T \to \infty$, we obtain \eqref{eq:localisation_rate}.

Finally, the $O_p(\delta_T)$ bound on the maximal localization error follows directly from \eqref{eq:localisation_rate}: for any $\varepsilon > 0$ there exists $T_0$ such that for all $T \ge T_0$,
\[
\Pr\Bigl(
  \max_{1 \le k \le K} \min_{0 \le j \le \widehat K} |\widehat\tau_j - \tau_k^\star|
  \le \delta_T
\Bigr)
\ge 1 - \varepsilon,
\]
which is exactly $\max_k \min_j |\widehat\tau_j - \tau_k^\star| = O_p(\delta_T)$.

This completes the proof.

\section{Computational complexity of KCPD}

The computational complexity of kernel change point detection (KCPD) combined with the PELT algorithm is well understood \citep{arlot-2019}. Let $n$ denote the number of observations and $c$ the cost of evaluating the kernel function. In the exact kernel setting, forming the $n\times n$ Gram matrix requires $O(n^2c)$ time and $O(n^2)$ memory. Given the Gram matrix, segment costs can be evaluated in $O(1)$ time via cumulative sums, and PELT achieves linear expected time under its standard pruning assumptions \citep{Killick01122012} (with a quadratic worst case). Consequently, in the exact-kernel setting the overall time and memory are typically dominated by Gram-matrix precomputation, i.e., $O(n^2c)$ time and $O(n^2)$ memory.

We use cosine similarity implemented as a dot product on unit-normalized embeddings (i.e., a linear kernel on normalized features). For the standard KCPD within-segment scatter cost
\[
C(s,e) \;=\; \sum_{t=s}^e k(t,t)\;-\;\frac{1}{e-s+1}\sum_{t=s}^e\sum_{u=s}^e k(t,u),
\]
the linear kernel yields the closed form
\[
C(s,e) \;=\; L \;-\; \frac{1}{L}\left\|\sum_{t=s}^e y_t\right\|_2^2,
\qquad L=e-s+1,
\]
since $k(t,u)=y_t^\top y_u$ and $\|y_t\|_2=1$. Precomputing prefix sums $P_t=\sum_{i=1}^t y_i$ allows evaluating $C(s,e)$ in $O(d)$ time using $P_e-P_{s-1}$, where $d$ is the embedding dimension, without forming the Gram matrix. This reduces memory to $O(nd)$ (or $O(d)$ if embeddings are streamed and only prefix sums are stored), and makes the PELT optimization close to linear in $n$ in practice, in addition to the one-pass embedding computation.

\section{Additional Experimental Results}\label{app:sims}
\begin{figure*}[h]
    \centering
    \begin{minipage}{0.48\linewidth}
        \centering
        \includegraphics[width=\linewidth]{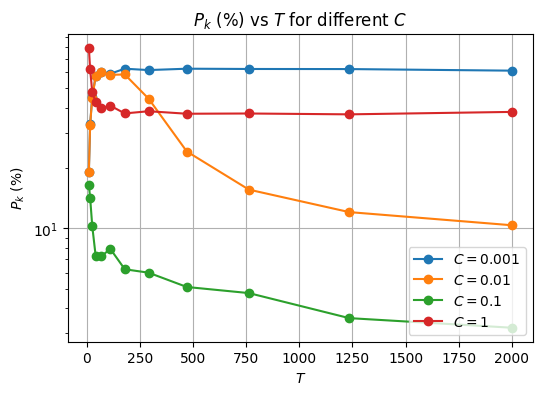}
        \caption{$P_k$ error (\%) versus sequence length $T$ for Embed-KCPD applied to synthetically generated short-range dependent text data with GPT-4.1, $m=20$, for multiple values of $C$ and sBERT embeddings.}
        \label{fig:sensitivity}
    \end{minipage}\hfill
    \begin{minipage}{0.48\linewidth}
        \centering
        \includegraphics[width=\linewidth]{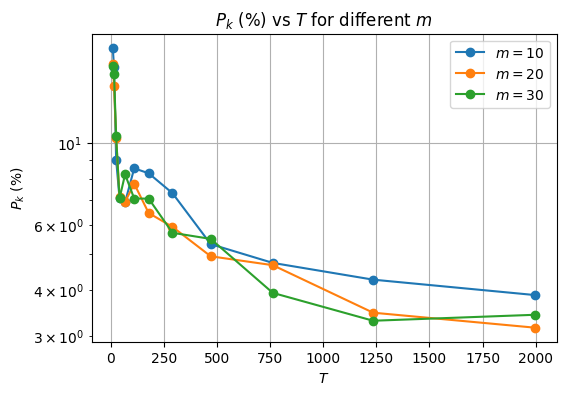}
        \caption{$P_k$ error (\%) versus sequence length $T$ for Embed-KCPD applied to synthetically generated short-range dependent text data with GPT-4.1, $C=0.1$, for multiple values of $m$ (number of sentences in LLM generation) and sBERT embeddings.}
        \label{fig:msensitivity}
    \end{minipage}
\end{figure*}

Figures~\ref{fig:sensitivity} and \ref{fig:msensitivity} indicate the effect of varying $C$ and $m$ on $P_k$.

\section{Experimental Details}\label{app:simulation}

\subsection{Statistics of Dataset}\label{ref:summary}
Here is the summary of all datasets we used in the experiments. Table~\ref{tab:stats_dataset} present the summary statistics for each dataset: total number of documents, number of segments per document, number of sentences per segment.
\begin{table}[h]
    \centering
    \caption{Statistics of Datasets in Our Experiments.}
    \begin{tabular}{lccc}
    \toprule
        \textbf{Dataset} & \textbf{Documents} & \textbf{Segments per Document} & \textbf{Sentences per Segment} \\
        \hline
        Choi (3-5)  & 100 & 10 & 4.0\\
        Choi (6-8)  & 100 & 10 & 7.0\\
        Choi (9-11) & 100 & 10 & 9.9\\
        Choi (3-11) & 400 & 10 & 7.0\\
        Wiki-300 & 300 & 7.6 & 26.0 \\
        Wiki-50 & 50 & 8.2 & 7.5 \\
        Elements & 118 & 7.7 & 2.9 \\
        arXiv & 20 & 9.5 & 7.1\\
        \bottomrule
    \end{tabular}
    
    \label{tab:stats_dataset}
\end{table}

\subsection{Implementation details}\label{app:implementation}

Embed-KCPD is implemented with the ruptures library \citep{truong2020selective}, using its kernel-based change-point implementation. We use ruptures'  median heuristic to set the bandwidth for the RBF Kernel. 

We compute text-embedding-3-small sentence representations using the OpenAI API. For the LLM-based experiment, we use GPT-4.1 via the same API; the total API cost for running all experiments is below \$20. All other embedding backbones are computed locally with the sentence-transformers library using the corresponding pretrained models.

For all baseline methods, we use the fine tuned hyperparameters from the original papers or from widely used public implementations. 

All code and implementation is available as supplementary materials.
\subsection{Optimal $C$ via Elbow Method}\label{app:optimal_c}
\begin{figure}
    \centering
    \includegraphics[width=\linewidth]{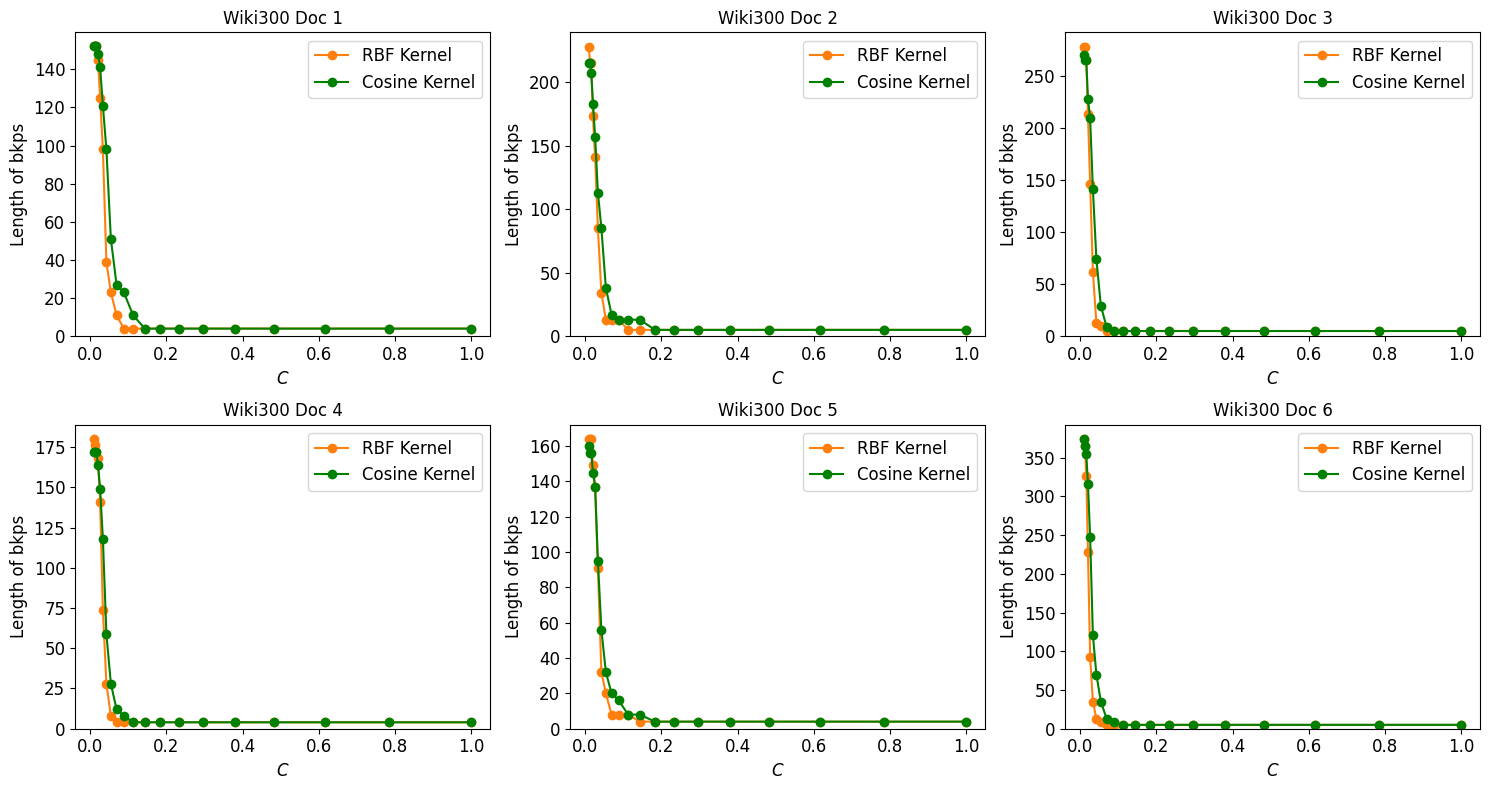}
    \caption{Sensitivity of the number of detected segments to the hyperparameter $C$ on Wiki-300.}
    \label{fig:c_wiki300}
\end{figure}

\begin{figure}
    \centering
    \includegraphics[width=\linewidth]{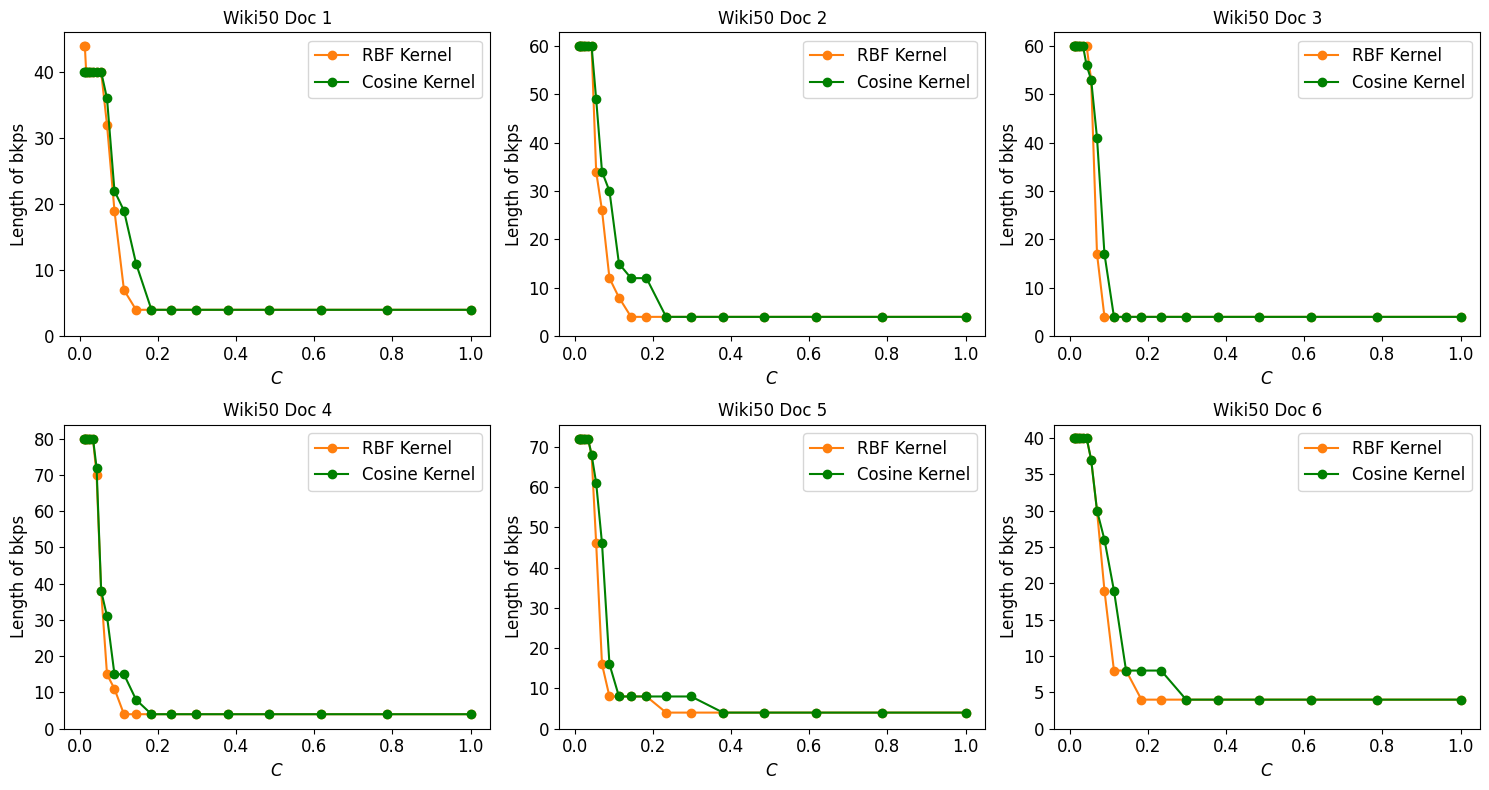}
    \caption{Sensitivity of the number of detected segments to the hyperparameter $C$ on Wiki-50.}
    \label{fig:c_wiki50}
\end{figure}

\begin{figure}
    \centering
    \includegraphics[width=\linewidth]{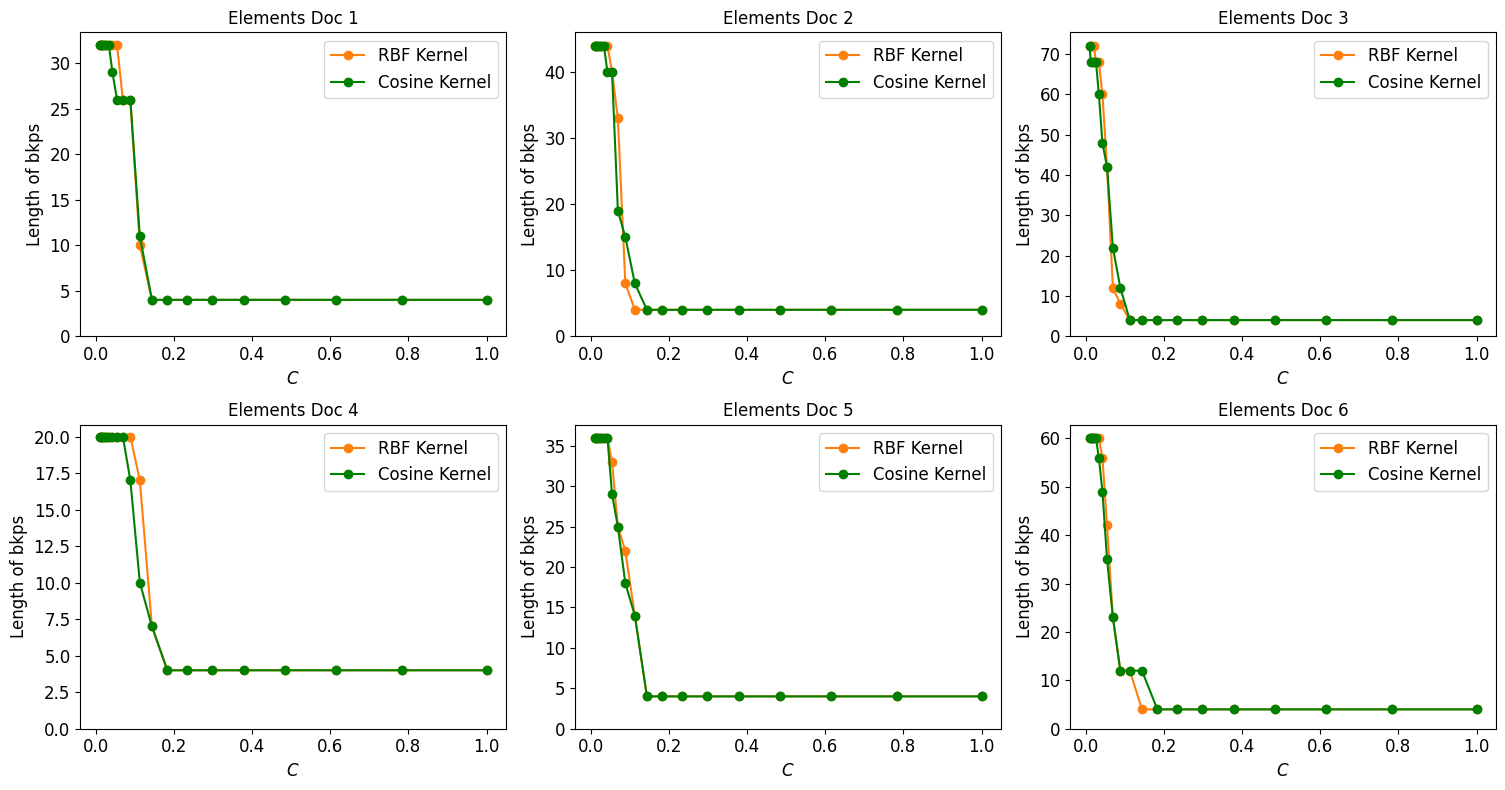}
    \caption{Sensitivity of the number of detected segments to the hyperparameter $C$ on Elements.}
    \label{fig:c_elements}
\end{figure}

\begin{figure}
    \centering
    \includegraphics[width=\linewidth]{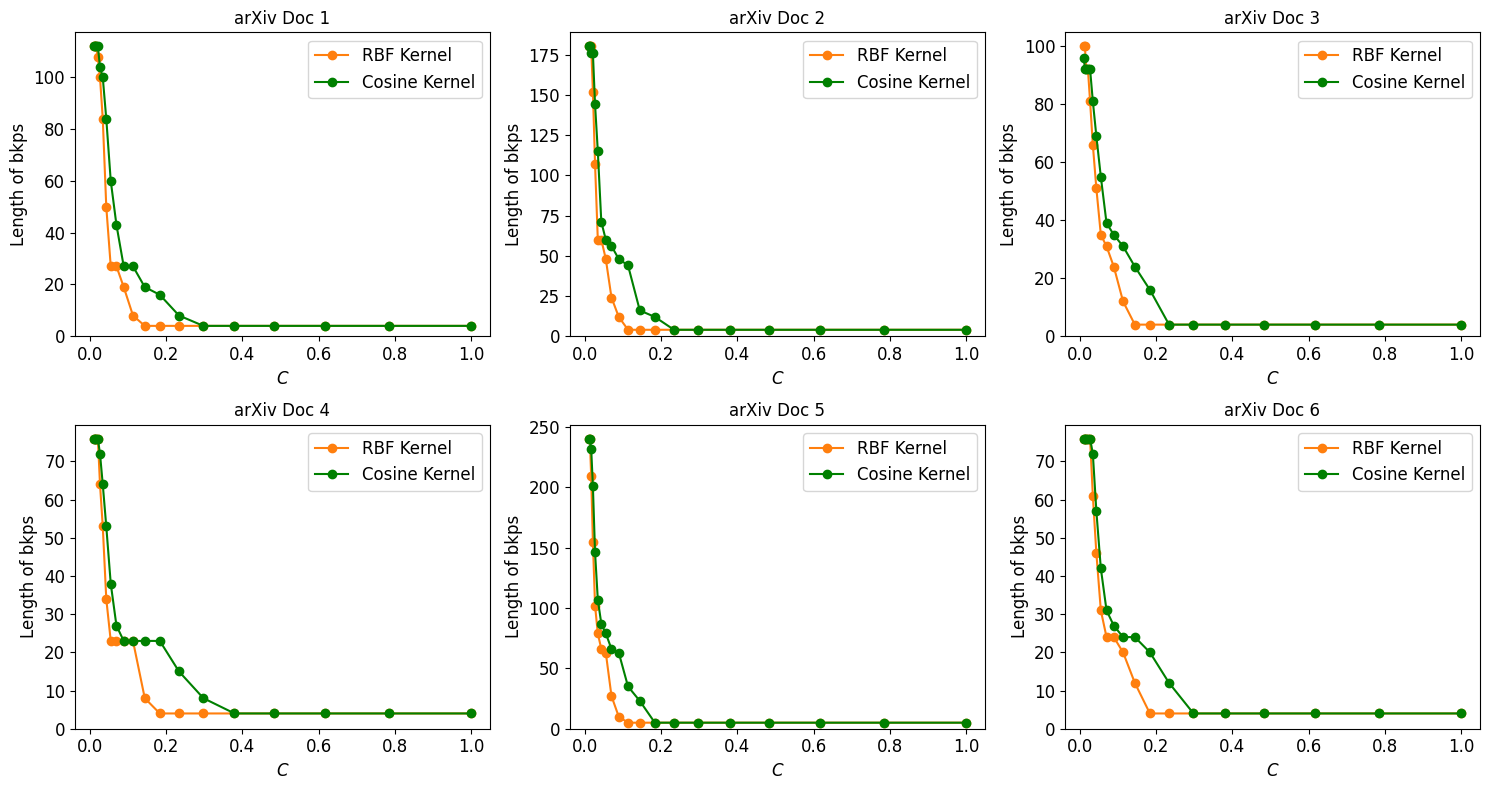}
    \caption{Sensitivity of the number of detected segments to the hyperparameter $C$ on arXiv.}
    \label{fig:c_arxiv}
\end{figure}

\begin{figure}
    \centering
    \includegraphics[width=\linewidth]{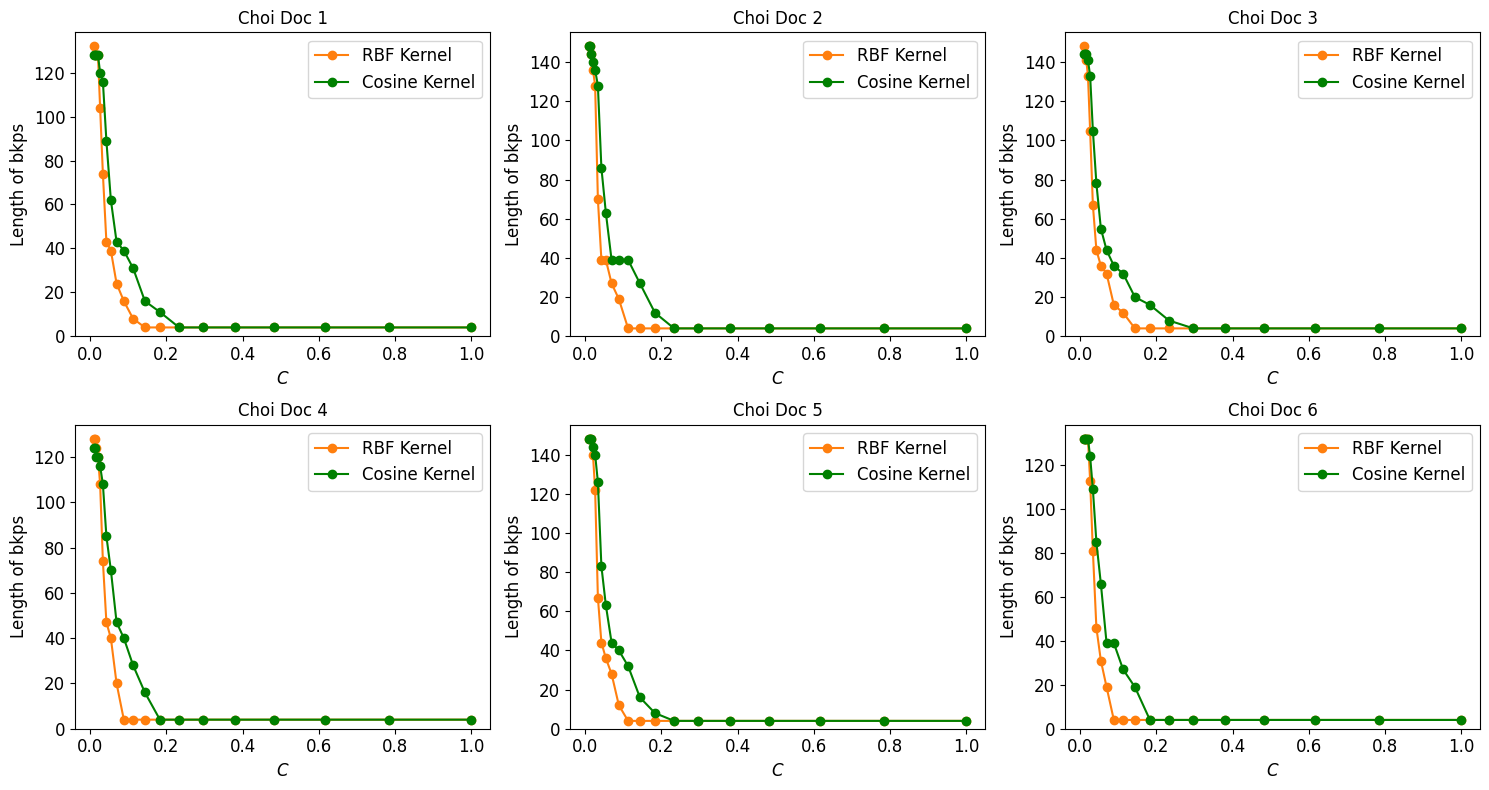}
    \caption{Sensitivity of the number of detected segments to the hyperparameter $C$ on Choi (3-11).}
    \label{fig:c_choi}
\end{figure}

For each dataset, we randomly sample 6 documents and, for each document, run Embed-KCPD over a small logarithmically spaced $C$ in the range $[10^{-2}, 10^0]$. The elbow point of the curve relating the number of detected change points to $C$ is selected per document, and the final $C$ is set to the average of these six values (see Fig.~\ref{fig:c_wiki300} for 6 documents from Wiki-300 datasets). Across datasets, the resulting elbow locations are highly consistent (see Figures~\ref{fig:c_wiki50}-\ref{fig:c_choi}). We therefore fix $C=0.06$ for the RBF kernel and $C=0.088$ for the cosine kernel across all experiments to 
ensure a fair, unsupervised comparison. Since \(\beta_T = C\sqrt{T\log T}\), the effective penalty adapts to sequence length.


\begin{figure}
    \centering
    \includegraphics[width=0.5\linewidth]{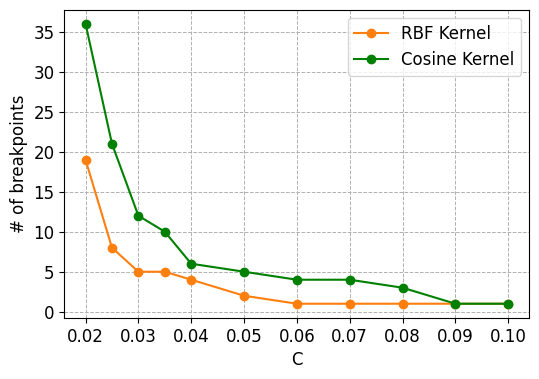}
    \caption{Sensitivity of the number of detected segments to the hyperparameter $C$ on Taylor Swift’s tweet stream.}
    \label{fig:c_taylor}
\vspace{-0.2cm}
\end{figure}

\begin{figure}[h]
\centering
 \includegraphics[width=0.95\linewidth]{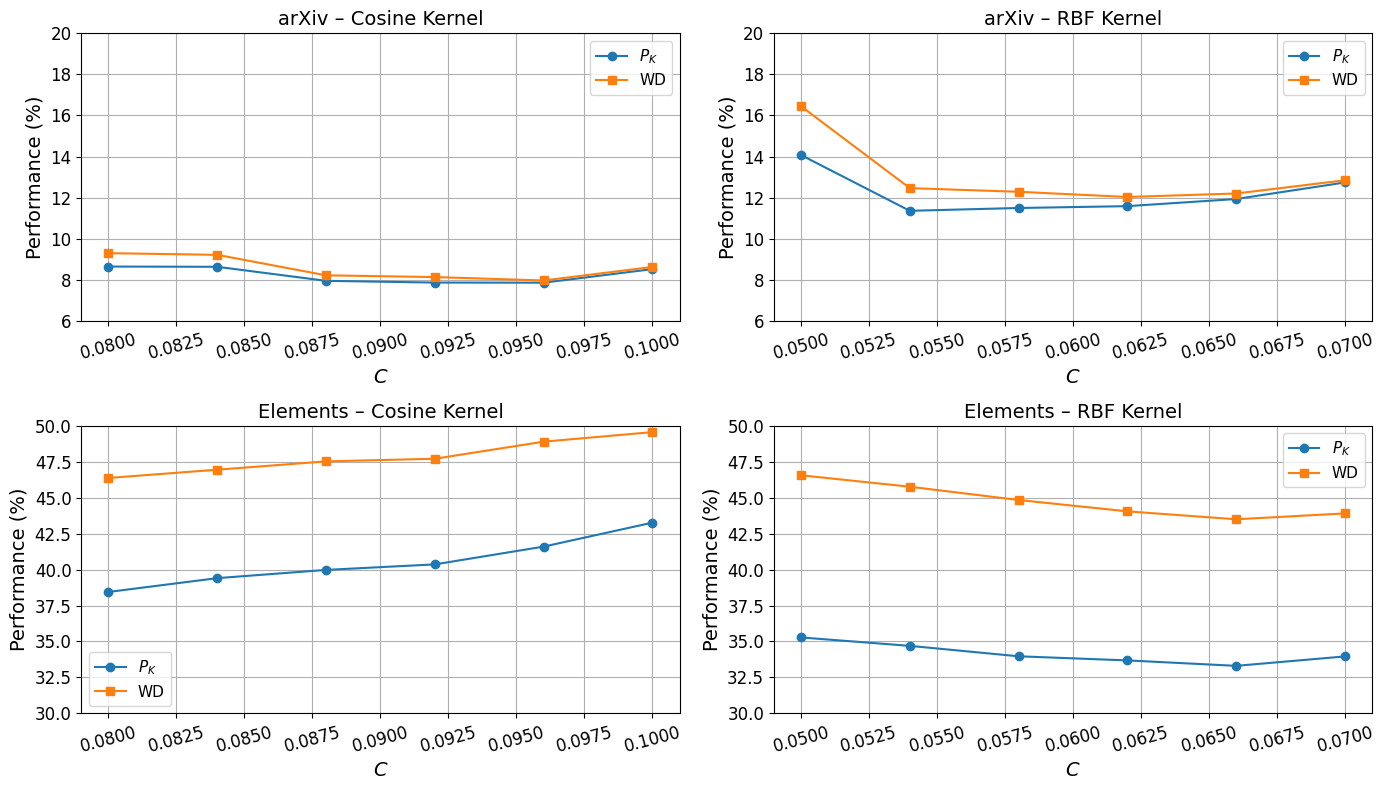} 
\caption{Sensitive of $C$ with cosine and RBF kernel on Elements and arXiv dataset.}
\label{fig:c_pk}
\end{figure}

\subsection{Sensitivity of $C$ on $P_k$ and WD} \label{app:C_pk}
To validate the robustness of our method with respect to $C$ around the identified sweet spots, $C = 0.088$ for the kCPD kernel and $C = 0.06$ for RBF kernel, we conduct a sensitivity analysis on arXiv and Elements datasets. As shown in Figure~\ref{fig:c_pk}, we vary $C$ within the range $[0.08, 0.10]$ for kCPD and $[0.05, 0.07]$ for RBF. Across these intervals, both the $P_k$ and WD metrics remain stable, indicating that performance is not sensitive to small perturbations of $C$ near the optimal region.

\subsection{$m$-dependent Data Generation}\label{app:m-dependent data}
Sentences are generated sequentially using GPT-4.1 with the fixed prompt: \textit{Give me one more sentence to naturally continue the text specific on [Topic]. Do not add any preamble just answer with one sentence. [Input Sentences].}

For a target sequence length $T$, the number of change points, \(K=\lceil 2\log T\rceil\), increases slowly with T. Candidate change-point locations are sampled uniformly without replacement from ${1, \dots, T}$, and converted into $K+1$ segment lengths via successive differences. No explicit minimum segment length constraint is imposed. Instead, segment lengths are controlled implicitly: as $T$ grows, the average segment length ${T}/{(K+1)}$ also grows, ensuring that segments become longer asymptotically, consistent with the minimum spacing requirement in Assumption~\ref{A4}.

\subsection{arXiv Dataset Generation}\label{ref:arxiv}
We construct new dataset based on the recent paper abstracts for text segmentation. The generation process is as follows:
\begin{itemize}
    \item Select the first 1000 papers from arXiv published after August 2025.
    \item Randomly sample 20 values between 5 and 20 to determine the number of unique abstracts per document.
    \item For each document, randomly select the corresponding number of abstracts, shuffle them to concatenate into a single text. Repeat this process 20 times to obtain 20 documents.
 
\end{itemize}

Full list of 1000 arXiv papers used to built this dataset part of the supplementary materials.

\section{Data Disclaimer}
We collected tweets from Taylor Swift’s official Twitter/X account (@taylorswift13) between January 2020 and May 2025, totaling approximately 400 posts. These tweets are public user-generated content, and our study only uses them for aggregate statistical analysis. In compliance with Twitter/X’s Terms of Service, we do not redistribute the dataset; instead, our paper reports only derived analyses. Code to extract these tweets using X API part of the supplementary materials.

\end{document}